\documentclass[journal]{IEEEtran}
\IEEEoverridecommandlockouts
\usepackage{amsmath,amssymb,amsfonts}
\usepackage{algorithmic}
\usepackage{graphicx}
\usepackage{textcomp}

\usepackage{booktabs}
\usepackage{subfig}
\usepackage[table,xcdraw]{xcolor}

\usepackage{comment}
\usepackage{todonotes}
\usepackage{orcidlink}
\usepackage[capitalise]{cleveref}
\usepackage{natbib}
\usepackage{multirow}
\usepackage{enumitem}

\newcommand{\ours}{M2CTS\,}

\newcommand{\cf}{\emph{cf.}~} 

\newif\ifanonym      
\anonymfalse        
\newif\ifarxiv      
\arxivtrue

\def\BibTeX{{\rm B\kern-.05em{\sc i\kern-.025em b}\kern-.08em
    T\kern-.1667em\lower.7ex\hbox{E}\kern-.125emX}}
\begin{document}


\title{Checkmating One, by Using Many: Combining Mixture of Experts with MCTS to Improve in Chess}

\ifanonym
\author{Authors anonymized}
\else
\author{\IEEEauthorblockN{Helfenstein Felix$^{\ast}$\thanks{$^{\ast}$These authors contributed equally.}}
\IEEEauthorblockA{\textit{Computer Science Department} \\
\textit{TU Darmstadt}\\
Darmstadt, Germany \\
{felix.helfenstein@stud.tu-darmstadt.de} \orcidlink{0000-0001-5681-532X}} \\
\and
\IEEEauthorblockN{Czech Johannes$^{*}$}
\IEEEauthorblockA{\textit{Artificial Intelligence and Machine Learning Lab} \\
\textit{TU Darmstadt}\\
Darmstadt, Germany \\
{johannes.czech@cs.tu-darmstadt.de} \orcidlink{0000-0002-9568-8965}} \\
\and
\IEEEauthorblockN{Blüml Jannis$^{*}$}
\IEEEauthorblockA{\textit{Artificial Intelligence and Machine Learning Lab, Hessian Center for Artificial Intelligence (hessian.AI)} \\
\textit{TU Darmstadt}\\
Darmstadt, Germany \\
blueml@cs.tu-darmstadt.de \orcidlink{0000-0002-9400-0946}} \\
\and
\IEEEauthorblockN{Eisel Max}
\IEEEauthorblockA{\textit{Computer Science Department} \\
\textit{TU Darmstadt}\\
Darmstadt, Germany \\
{max.eisel@stud.tu-darmstadt.de} \orcidlink{0009-0004-9082-9871}} \\
\and
\IEEEauthorblockN{Kersting Kristian}
\IEEEauthorblockA{\textit{Artificial Intelligence and Machine Learning Lab, Hessian Center for Artificial Intelligence (hessian.AI), Centre for Cognitive Science, German Research Center for Artificial Intelligence (DFKI)} \\
\textit{TU Darmstadt}\\
Darmstadt, Germany \\
kersting@cs.tu-darmstadt.de \orcidlink{0000-0002-2873-9152}}
}
\fi

\maketitle

\begin{abstract}
In games like chess, strategy evolves dramatically across distinct phases — the opening, middlegame, and endgame each demand different forms of reasoning and decision-making. Yet, many modern chess engines rely on a single neural network to play the entire game uniformly, often missing opportunities to specialize.
In this work, we introduce M2CTS, a modular framework that combines Mixture of Experts with Monte Carlo Tree Search to adapt strategy dynamically based on game phase.
We explore three different methods for training the neural networks: Separated Learning, Staged Learning and Weighted Learning. By routing decisions through specialized neural networks trained for each phase, M2CTS improves both computational efficiency and playing strength.
In experiments on chess, M2CTS achieves up to +122 Elo over standard single-model baselines and shows promising generalization to multi-agent domains such as Pommerman. These results highlight how modular, phase-aware systems can better align with the structured nature of games and move us closer to human-like behavior in dividing a problem into many smaller units.
\end{abstract}



\section{Introduction}
Chess has long stood as a benchmark for artificial intelligence — not just because of its complexity but also because of its potential to reveal the limits and capabilities of reasoning systems. Its rich structure, long-term dependencies, and diverse tactical and strategic demands make it a valuable testbed for learning and planning algorithms. Systems like AlphaZero~\citep{AlphaZero} have achieved remarkable success by combining deep reinforcement learning with Monte Carlo Tree Search (MCTS), reaching superhuman performance through self-play alone.

However, these systems treat the game as a single, uniform problem, relying on one neural network to make decisions across all stages — from the structured precision of the opening, to the tactical depth of the middlegame, to the calculation-heavy endgame. While this simplifies architecture and training, it may overlook the reality that each phase poses fundamentally different challenges. In contrast to the early game’s reliance on well-established theory and position development, the endgame often requires fine-grained calculation under constrained material.

One concrete issue that arises from this design is data imbalance. Because most training positions come from the middlegame, models tend to overfit to this phase, often underperforming in openings and endgames, where data is scarcer and strategic demands differ~\citep{McIlroy-Young0K20, McIlroy-YoungW022, palsson2023unveiling}. This bias not only affects evaluation accuracy but can also limit the agent’s ability to discover effective play in underrepresented regions of the game tree.\\

Despite clear structural differences between the opening, middlegame, and endgame, state-of-the-art engines such as AlphaZero~\citep{AlphaZero}, MuZero~\citep{Schrittwieser2020MuZero}, and Stockfish NNUE~\citep{nasu2018efficiently} continue to rely on monolithic architectures trained uniformly across all game phases. This uniform treatment simplifies training but raises an important question: Can we build phase-aware architectures — models that adapt their internal reasoning to match the shifting demands of the task~\citep{AgarwalaDJPSWZ21, Dobre2017combining}? 

\begin{figure*}[t]
    \centering
    \ifarxiv
    \includegraphics[width=0.9\textwidth]{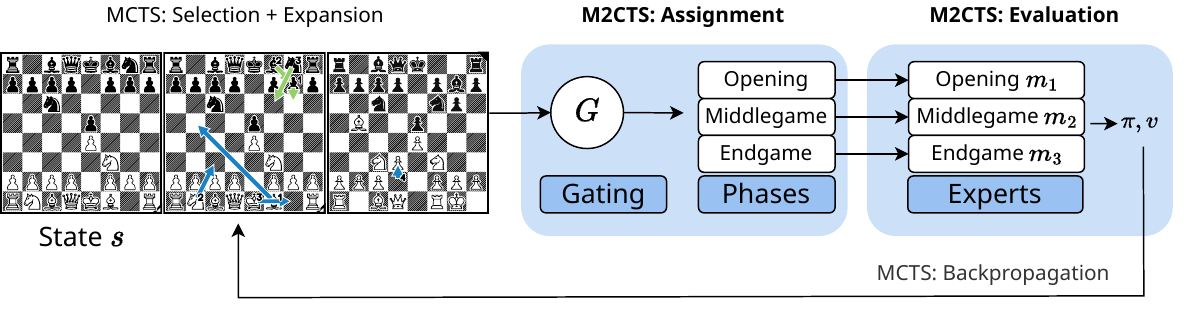}
    \else
    \includegraphics[width=0.9\textwidth]{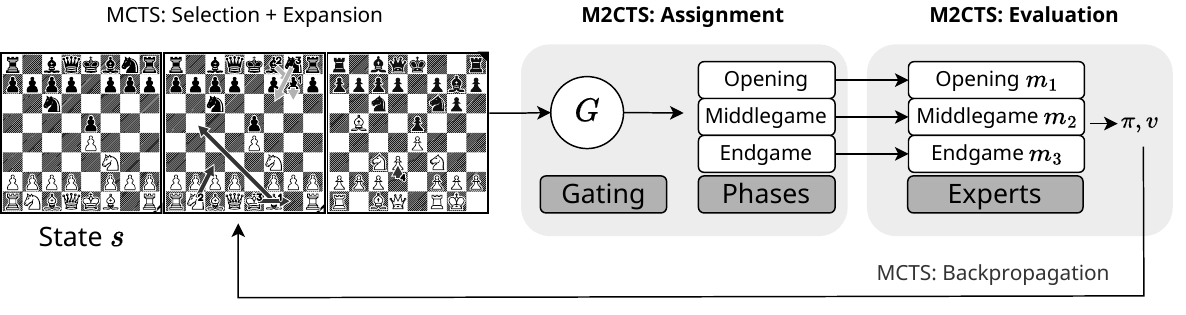}
    \fi
    \caption{{\ours incorporates MoE-enhanced MCTS through phase-based model selection.} \ours splits the original simulation phase into two distinct phases. First, we need to categorizing game states into phases using a gating mechanism in the \textit{Assignment phase}, then selecting models for evaluation and backpropagation (\textit{Evaluation phase}).}
    \label{fig:mcts_phases_diagram}
\end{figure*}

To explore this direction, we propose M2CTS, a modular framework that integrates Mixture of Experts (MoE)~\citep{JacobsJNH91} into an AlphaZero-style MCTS engine. As illustrated in \cref{fig:mcts_phases_diagram}, M2CTS dynamically selects between specialized neural networks — each trained for a specific phase — to evaluate game states during search.
By explicitly modeling the temporal structure of the game, M2CTS improves both computational efficiency and strategic accuracy. In chess, it achieves up to +122 Elo compared to standard single-model baselines, and demonstrates promising generalization to multi-agent domains such as Pommerman. These results suggest that even coarse-grained phase segmentation can provide meaningful gains, supporting the case for modular architectures in long-horizon, structured decision-making tasks. \\

\noindent To summarize, our contributions are as follows: 
\begin{enumerate}[leftmargin=20pt,itemsep=2pt,parsep=2pt,topsep=1pt,partopsep=2pt]
\item[(i)] We propose \ours, integrating MoE into MCTS frameworks like AlphaZero. 
\item[(ii)] We show that a simple separated learning strategy is as effective, or even more so, than complex alternatives when sufficient data is available. 
\item[(iii)] Our modular architecture can outperform comparable monolithic approaches in practice.
\end{enumerate}

We first describe \ours, covering phase-specific MoE design and four training strategies. Next, we evaluate its performance against traditional single-network models. Lastly, we discuss related work, implications, and limitations.

\section{M2CTS: Modular Search with Phase-Specific Neural Experts}
\label{sec:Methodology}
M2CTS integrates the MoE model into MCTS to optimize strategic decision-making in chess engines. By leveraging phase-specific expert models, M2CTS dynamically adapts its strategy as the game progresses. In this work, we will use chess as an example for explaining the methodology and as an environment for our experimental section. This section explains the gating mechanism that categorizes game states and activates the appropriate expert model and give insight in the training strategies for our experts. 



\subsection{Game Phase Definitions as Gating Mechanism}
\label{sec:gamephases}

\begin{figure}[t]
    \centering
    \ifarxiv
     \includegraphics[width=0.9\linewidth]{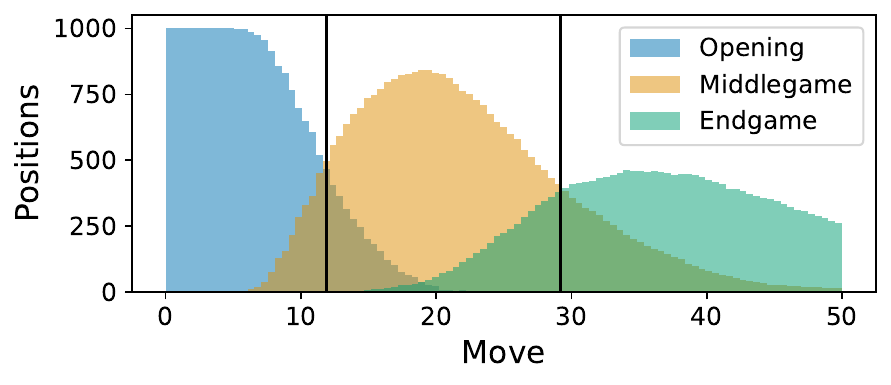}
    \else
     \includegraphics[width=0.9\linewidth]{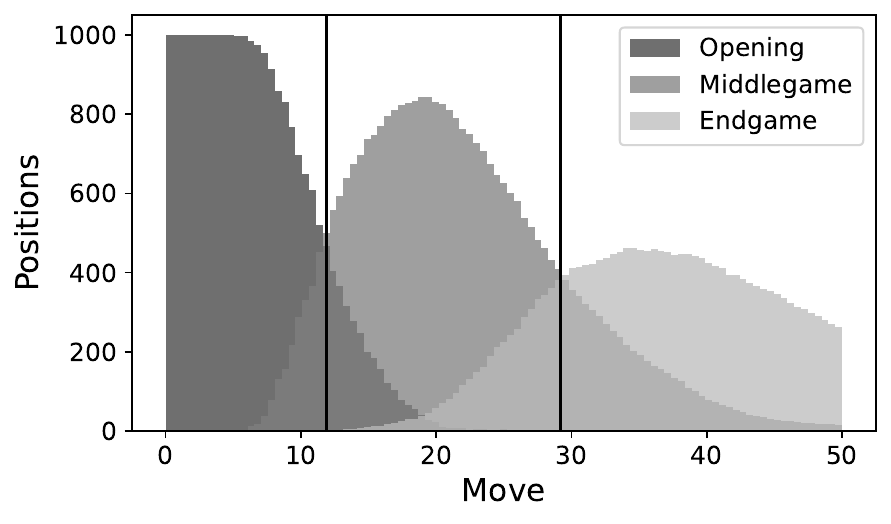}
     \fi
        \caption{{Splitting the test data into 3 phases: \textit{opening, middlegame} and \textit{endgame}.} A detailed distribution of positions across the three game phases based on the Lichess phase definition, aggregated by move number. The average starting moves of the middlegame ($11.89\pm2.97$) and the endgame ($29.15\pm7.44$) are marked by vertical lines.}
    \label{fig:phases}
\end{figure}

The MoE framework consists of multiple specialized neural networks, or “expert networks,” each designed to handle a specific game phase, \cf $m_1,...m_n$ in \cref{fig:mcts_phases_diagram}. 

These experts are guided by a ``gating mechanism'' that evaluates the current game state and selects the appropriate expert based on predefined phase criteria, as illustrated in \cref{fig:mcts_phases_diagram}, where it is displayed as $G$. In this work, coming from chess, we define these game phases as the opening, middlegame, and endgame, each with its own unique strategic demands.

To classify the game phase, we utilize the well-defined phase definitions from Lichess\footnote{\url{https://lichess.org/}, accessed 2024-05-22}, a widely-used open-source chess platform. These definitions consider factors such as piece count, board complexity, and back-rank vulnerability to determine the phase of the game. The gating mechanism analyzes the board state in real-time and activates the expert model that best suits the current phase, ensuring that the chess engine adapts its strategy dynamically. This approach provides a holistic perspective on the game's progression, enabling us to determine the phase of any given game state independently while maintaining the Markov property, crucial for accurate decision-making without historical dependencies. Additionally, this intuitive method accommodates transitions between phases based on the current board state, ensuring strategic adaptability throughout the game. For a detailed exploration of the Lichess phase definitions, please refer to the appendix. The distribution of game phases using these criteria is visualized in \cref{fig:phases}.
There exist several combination strategies for MoE models, with averaging and the max operator being the most common~\citep{Yang2004Combining}: \begin{align}
y = \sum_i^n G_i(x)m_i(x) \quad \textrm{or} \quad y = \max_{G_i(x)}(m_i(x)) 
\end{align} 
For this work, we adopt the max operator for its clarity in selecting a single, relevant phase, thus eliminating the possibility of ambiguous phase assignments. This method of pre-defined input segmentation streamlines the learning process by ensuring each expert focuses solely on its designated phase. Additionally, it brings significant computational efficiencies, as it requires inference from only one expert model $m_i$ at a time. However, this approach assumes accurate and classifiable game phases. While training a gating network could potentially improve adaptability by learning from the data, we have chosen not to pursue this due to the added complexity and the risk of overfitting. Instead, we prioritize a straightforward, rule-based gating mechanism that effectively leverages established phase definitions to enhance performance without the uncertainties associated with training.

\subsection{Phase Definitions and Working in Batches.} 
The ideal scenario involves querying always the appropriate network, aligned with the current phase of the MCTS position. However, this approach becomes impractical for MCTS with batch sizes larger than one, which are commonly used to improve GPU utilization. Larger batch sizes necessitate buffering multiple positions before neural network inference, leading to reduced node statistics updates and can end in querying multiple networks since the batch has not one isolated phase anymore.

To tackle this issue, we devised a strategic approach for handling larger batches to reduce the computational load again. When faced with a batch size greater than one, we analyze all positions within the batch to determine the dominant game phase. The network of experts aligned with this dominant phase is then applied to the entire batch. While this adaptation may sometimes process instances with a less-than-ideal expert network, such instances are expected mainly during phase transitions. In such scenarios, even if positions are incorrectly processed, they remain close to their appropriate phase, minimizing any compromise in prediction accuracy. This modification is crucial for optimizing GPU efficiency and effectively managing larger batch sizes, important aspects in computation chess.

\begin{figure*}[t]
    \centering
    \ifarxiv
    \includegraphics[width=\linewidth]{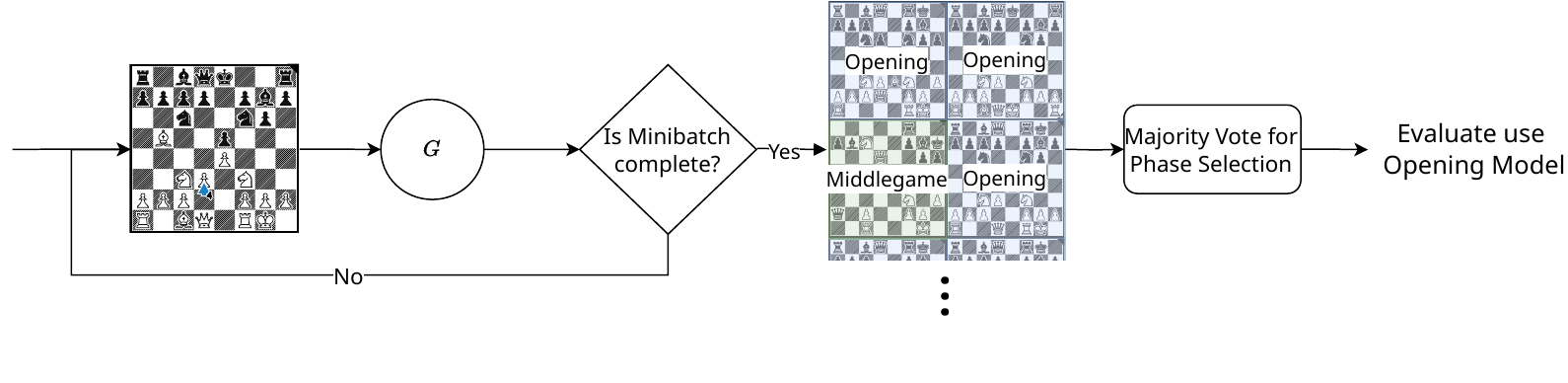}
    \else
    \includegraphics[width=\linewidth]{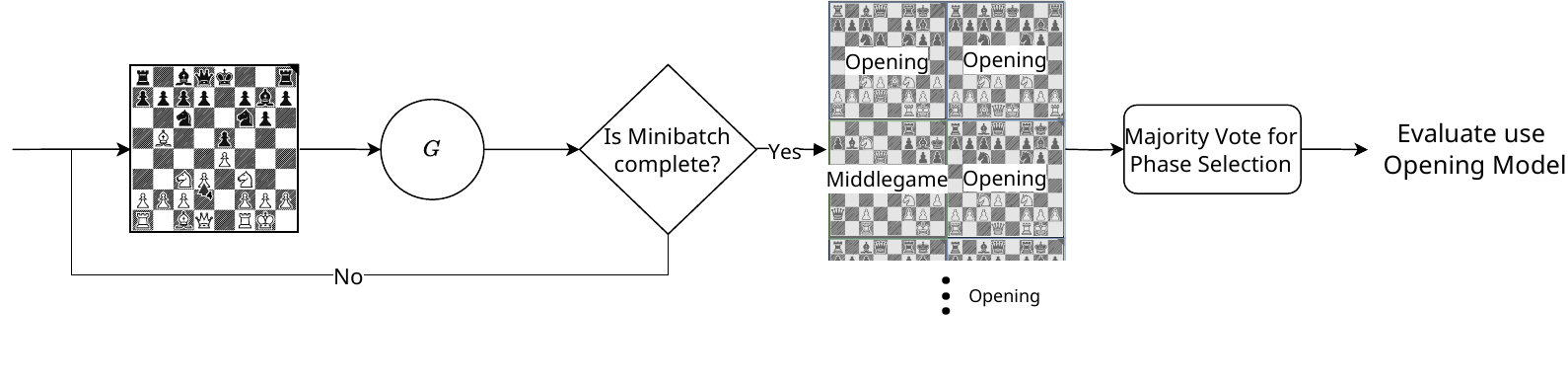}
    \fi
    \caption{{MCTS in combination with MoE can be used for batch sizes greater than 1.} Each sample is assigned to its phase while allocating a new mini-batch during the MCTS search. When a mini-batch is complete, majority vote selection is used to select the model that contains the most samples of its associated phase. This model is then used to evaluate all samples from this mini-batch ensuring efficient and phase-focused processing.}
    \label{fig:batches}
\end{figure*}

\subsection{Training Phase-Specific Experts}
\label{sec:trainingmethods}
Training an expert for a specific game phase can be as easy as classifying the data in a first step and splitting the training set into \(n\) parts. However, ensuring that each expert captures the nuances and transitions between phases could require more than just isolated training, like a comprehensive understanding of how the phases interact and evolve throughout the game. For this, we considered three distinct training methodologies T1--T3.
\begin{description}
\item{\textbf{T1 -- Separated Learning.}} This approach focuses on training each of the three expert networks exclusively on positions relevant to their corresponding game phases. By restricting the input space to specific phases, we facilitate focused expertise development, which in turn bolsters parameter counts without sacrificing inference speed. This method's simplicity allows parallelized training across experts, leading to significant reductions in training time. Although it ensures in-depth phase-specific knowledge, its drawback lies in potential challenges with transition nuances between phases. However, the method's efficiency and specificity maintain training times comparable to single-network approaches.

\item{\textbf{T2 -- Staged Learning.}} In this setup, knowledge is progressively transferred between phases, allowing earlier-stage learning to inform later-phase behavior. This can be done either by training sequentially on phase-specific data or by fine-tuning from a model trained on the full game. While this promotes broader understanding and aligns with transfer learning principles, it can also introduce challenges such as increased training time and the risk of forgetting earlier phase-specific patterns.

\item{\textbf{T3 -- Weighted Learning.}} This strategy emphasizes different sample weights in the loss function based on their game phase during training. For balancing, we employ 
\begin{align} L_{\text{weighted}} = \frac{1}{B} \sum_{n=1}^{B}w_n L(x_n) \end{align} where, $B$ is the batch size, $w_n$ the weight of a specific sample, and $L(x_n)$ the regular sample loss. To prioritize the samples from the actual phase, we introduce the parameter \(a\geq 1.0\), which balances the weights of the actual phase ($w_{\text {main}}$) and the ones of all other samples ($w_{\text {other}}$): $w_{\text {main}}=a \cdot w_{\text {other}}$. This approach ensures a consistent magnitude of loss across training and evaluation by normalizing the sample weights to maintain an average sample weight close to 1.0. Notably, the test set remains unweighted for fair performance comparison.
\end{description}
Each of these approaches is designed to enhance the MoE model's performance in MCTS. The detailed configurations and outcomes of these training strategies are further elaborated in the experimental section. 

\section{Experimental Evaluation}
\label{sec:experiments}

We empirically evaluate \ours{} across a range of settings to assess its effectiveness. Our experiments are guided by the following research questions:
\begin{enumerate}[leftmargin=25pt,itemsep=2pt,parsep=2pt,topsep=1pt,partopsep=2pt]
    \item[\textbf{(Q1)}] {Does M2CTS improve in performance over classical MCTS?}
    \item[\textbf{(Q2)}] {How do training strategies, expert composition, and gating mechanisms affect the performance and efficiency of M2CTS?}    
    \item[\textbf{(Q3)}] {Can M2CTS generalize beyond chess?}
\end{enumerate}
 
\subsection{Experimental Setup.}
\label{sec:setup}
We evaluate \ours{} under a range of controlled settings. During training, the model is evaluated every 500 iterations on both a dedicated validation set and all four test subsets (opening, middlegame, endgame, and no-phase). We save checkpoints whenever the current evaluation loss improves over the best previously recorded value. The final model is selected from the last such checkpoint, representing the best validation performance.

The backbone of our neural network architecture is RISEv3.3, a mobile convolutional network adapted from RISEv2~\citep{Czech2020CrazyAra}. It incorporates residual blocks and dual output heads for value and policy prediction, and is further enhanced with $5 \times 5$ convolutions and Efficient Channel Attention (ECA) modules~\citep{wang2020eca}. A detailed architecture diagram and explanation are provided in the appendix.

We adopt the input and output encoding from \citet{czech2024representation}. The input consists of a stack of spatial planes encoding piece positions, the previous eight moves, castling rights, material balance, and auxiliary metadata. Our framework builds on the ClassicAra variant of CrazyAra, designed specifically for classical chess.

We use training and test data derived from the KingBase Lite 2019 database\footnote{\url{https://archive.org/details/KingBaseLite2019}, accessed 2024-05-22}, which contains over one million games played by players rated 2200 Elo or higher. Games with fewer than five moves are excluded to filter out pre-arranged draws or corrupted records.

To improve training stability, we implement a loss spike recovery mechanism. If the validation loss increases sharply (by a factor of $s = 1.5$ or more), the model reverts to the previous best checkpoint. This helps prevent catastrophic performance drops and promotes steady convergence.

Our main evaluation consists of a 1000-game match using AlphaZero-style MCTS. We report Elo ratings for each configuration and compute 95\% confidence intervals using the method from the Cutechess framework\footnote{\url{https://github.com/cutechess/cutechess}, accessed 2024-05-22}.

Hyperparameters and train/test splits follow those used in ClassicAra, which has demonstrated competitive results in tournament settings. Full configuration details are available in the appendix. Training was conducted on 1–2 NVIDIA A100 GPUs; full runtime and resource usage details are reported in the appendix.

\begin{table*}[t]
\centering
\begin{tabular}{lrrrrrr}
\toprule
                                    & \multicolumn{6}{c}{\textbf{Batch Size}}                                                                                                                                                                                                                                                                                    \\ \cmidrule(lr){2-7} 
\multirow{-2}{*}{\textbf{Elo Gain vs. classical MCTS}} & \multicolumn{1}{c}{\textbf{1}}                     & \multicolumn{1}{c}{\textbf{8}}                     & \multicolumn{1}{c}{\textbf{16}}                    & \multicolumn{1}{c}{\textbf{32}}                    & \multicolumn{1}{c}{\textbf{64}}                    & \textbf{Average}                                  \\
\midrule
\textbf{Separated Learning} & 106.89 $\bullet$ & 123.45 $\bullet$ & 126.81 $\bullet$ & 124.36 $\bullet$ & {\textbf{129.49}} $\bullet$ & 122.20 $\bullet$\\ 
\textbf{Staged Learning} & 111.68 $\bullet$ & 120.81 $\bullet$ & 122.49 $\bullet$ & 125.00 $\bullet$ & {\textbf{125.55}} $\bullet$ & 121.11 $\bullet$ \\
\textbf{Weighted Learning ($a=4$)} & {\textbf{25.61}} $\bullet$ & 23.20 $\bullet$ & 23.52 $\bullet$ & 22.28 $\bullet$ & 21.31 $\bullet$ & 23.18 $\bullet$ \\
\textbf{Weighted Learning ($a=10$)} & 50.41 $\bullet$ & 52.42 $\bullet$ & 56.33 $\bullet$ & {\textbf{63.55}} $\bullet$ & 56.51 $\bullet$ & 55.84 $\bullet$ \\
\bottomrule
\end{tabular}
\caption{{M2CTS outperforms ($\bullet$) our baseline ``one-for-all'' MCTS approach in direct comparison.} This table presents the relative Elo gains achieved by our different training approaches over our baseline model across different batch sizes. The rightmost column aggregates the average Elo gain across all 55 experiments conducted for each method. A higher Elo value correspond to a higher difference in playing strength in favor of M2CTS over MCTS. The highest Elo gain is marked \textbf{bold}. A more detailed version can be found in the appendix.}
\label{tab:elo_overview}
\end{table*}

\begin{figure}[tb]
    \centering
      \ifarxiv
      \includegraphics[width=0.9\linewidth]{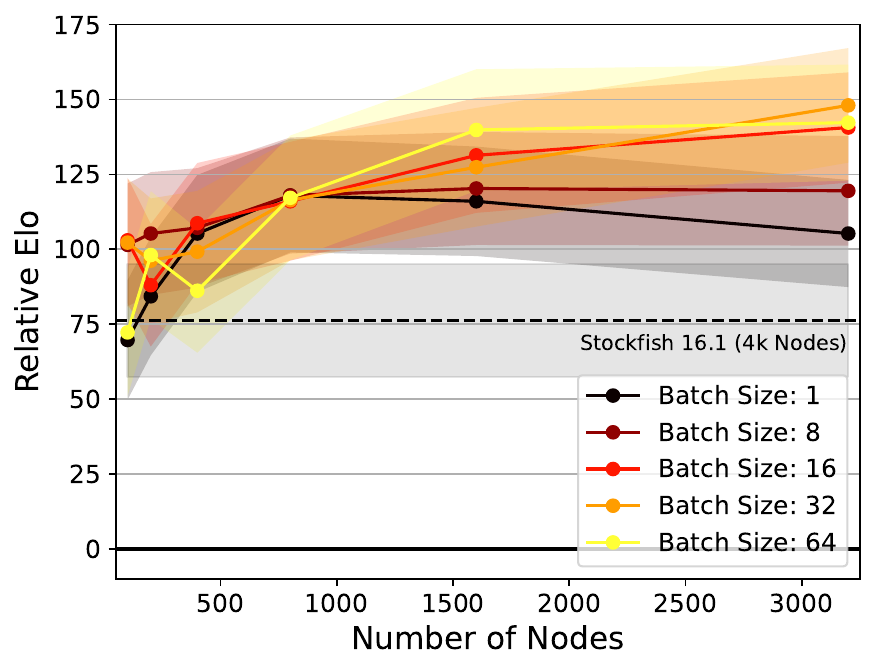}
      \else
      \includegraphics[width=0.9\linewidth]{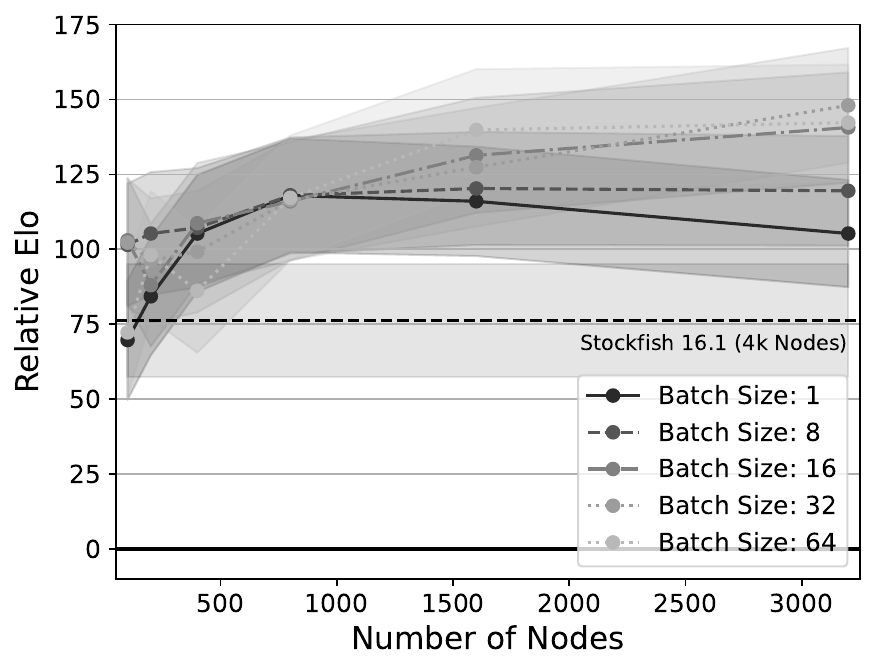}
      \fi
      \label{fig:sep_nodes}
    \caption{{M2CTS outperforms standard MCTS.} Utilizing the separated learning strategy outlined in Section~\ref{sec:trainingmethods}, achieves up to 150 Elo points more than our 'one-for-all' baseline. This increase in performance was consistently observed across a range of experiments, involving various batch sizes, and different tree complexities, measured in terms of nodes per tree.
    The dashed line serves as a reference point and represents the relative Elo of Stockfish 16.1 limited to 4000 nodes compared to the 'one-for-all' baseline with 3200 nodes.
    }
    \label{fig:sep_elo}
\end{figure}

\subsection{Performance Evaluation (Q1).}
\textbf{M2CTS outperforms classical MCTS.} Elo is a widely used metric for measuring relative playing strength in chess, with top engines like Stockfish, Leela, and ClassicAra reaching ratings above 3300 in competitive settings.

In our experiments, we trained phase-specific expert models using several learning strategies, introduced in \cref{sec:Methodology} and evaluated their combined performance within the M2CTS framework. As shown in \cref{tab:elo_overview}, \cref{tab:short_loss}, and \cref{fig:sep_elo}, M2CTS consistently improves over the single-model MCTS baseline. On average, our modular approach yields gains between 55 and 122 Elo points, depending on the training setup, with separated and staged learning performing best.

While M2CTS does not yet match the performance of elite engines like Stockfish (\cf \cref{tab:stockfish}), it demonstrates substantial progress over traditional MCTS, highlighting the benefit of incorporating game-phase specialization. These results suggest that integrating modular, context-aware strategies into high-performance engines could be a promising direction for future work. \\

\textbf{M2CTS achieves higher Elo than MCTS in reinforcement learning from random initialization.}
We evaluate \ours{} in a reinforcement learning (RL) setting where models start from random initialization and improve through self-play. We compare three configurations: (1) a classical MCTS baseline using a single network, (2) \ours{} with separated learning, and (3) \ours{} with staged learning. All models are trained for 10 update cycles using a shared replay buffer and fixed training budget. Each experiment is repeated across three random seeds. Full training details are provided in the appendix.
As shown in Figure~\ref{fig:rl_moe_tourney}, the staged learning variant of \ours{} achieves the highest Elo, outperforming the MCTS baseline by over 100 Elo points after training. In contrast, \ours{} with separated learning underperforms relative to the baseline. These findings suggest that when learning from scratch, a staged curriculum — in which a general model is first trained on the full dataset and subsequently fine-tuned for each phase — offers a more robust foundation for phase-aware decision-making.

\begin{figure}[tb]
    \centering
    \ifarxiv
      \includegraphics[width=0.9\linewidth]{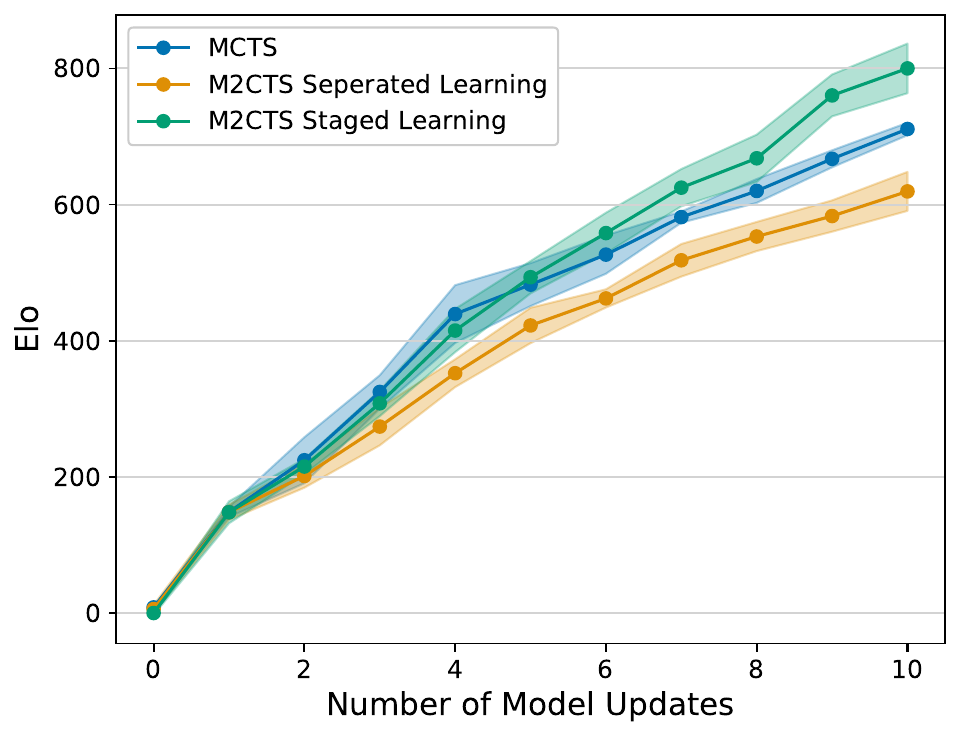}
    \else
      \includegraphics[width=0.9\linewidth]{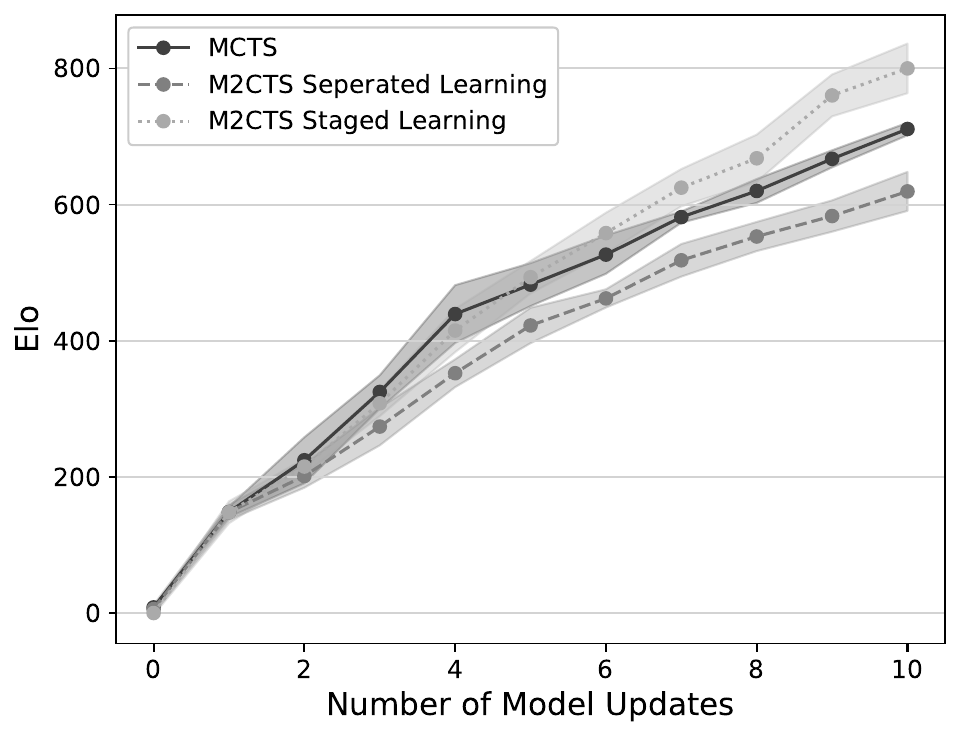}
      \fi
    \caption{\ours using staged learning outperforms MCTS during selfplay when starting from random weight initialization by more than 100 Elo points after 10 model updates.
    \ours with separated learning lacks behind by about 100 Elo after 10 model updates.
    }
    \label{fig:rl_moe_tourney}
\end{figure}

\subsection{Dissecting Architectural and Expert Design (Q2).}
\textbf{More complex training strategies do not outperform simple dataset splitting.}
We compare three expert training strategies: (i) simple dataset splitting (separated learning), (ii) staged learning, and (iii) weighted learning across all phases. In the supervised setting, separated learning — where each expert is trained only on data from its respective phase — consistently achieves the highest Elo (see \cref{tab:elo_overview}). This may stem from reduced task interference and more stable optimization when training is isolated per phase.

Despite their greater complexity, neither staged nor weighted learning outperforms this simpler baseline in the supervised setup. Only in the reinforcement learning setting, where much less data is available for each model update, does staged learning provide a clear advantage, outperforming separated learning by roughly 200 Elo points.

In weighted learning, increasing the specialization factor ($a = 10$ vs.\ $a = 4$) yields better performance. This suggests that emphasizing phase-relevant data may help bridge the gap — but comes at a cost. Weighted learning requires processing the full dataset for each expert, effectively tripling training time compared to separated learning , making it less sample-efficient and more resource-intensive.

Based on our experiments, we find separated learning to be the most effective and efficient training method in supervised scenarios with a lot of training data, and do not recommend weighted learning due to its inferior performance and higher cost. \\

\begin{figure}[tb]
    \centering
      \ifarxiv
      \includegraphics[width=0.9\linewidth]{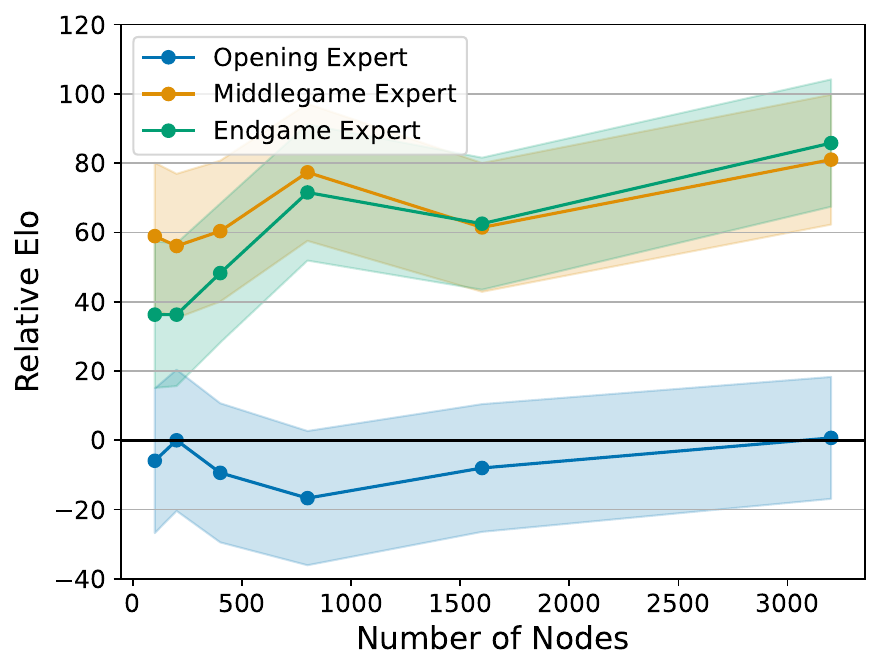}
      \else
      \includegraphics[width=0.9\linewidth]{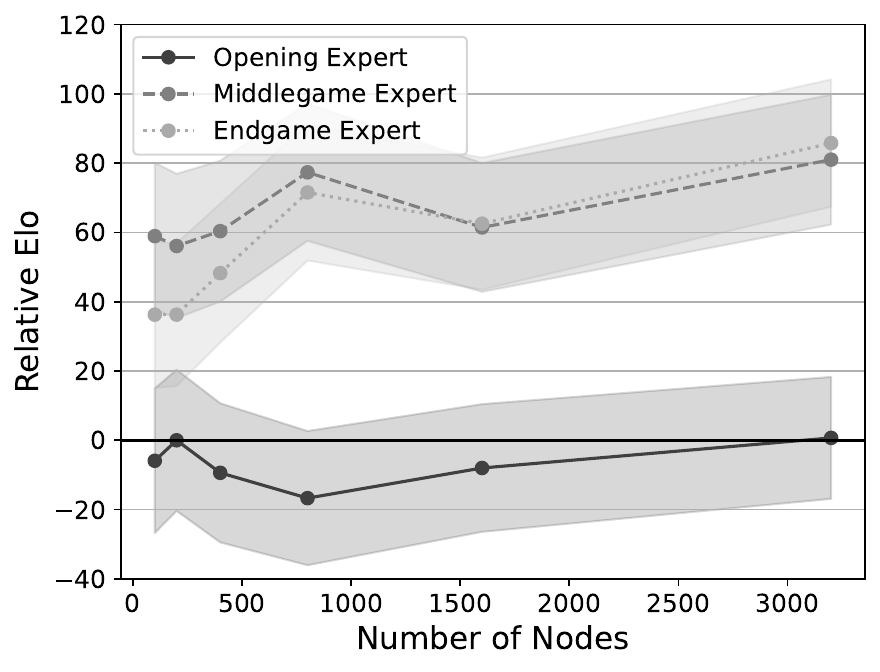}
      \fi
    \caption{{Expert models are especially helpful in middle- and endgame.} This figure shows the relative Elo gains achieved by selectively deploying individual phase experts within Monte-Carlo Tree Search (MCTS), compared to using a baseline network for all phases. The experts, derived from our separated learning approach, exhibit notable strengths in both the middlegame and endgame phases of chess. The experiments were conducted using a consistent batch size of 64. 
    }
    \label{fig:specific_elo}
\end{figure}

\textbf{Middlegame and endgame experts contribute most to performance, while the opening expert is limited by data constraints.}
To assess the individual impact of each phase-specific expert in \ours{}, we conducted ablation tournaments where only one expert model was active for its corresponding phase, while a baseline model was used for all others. As shown in \cref{fig:specific_elo}, both the middlegame and endgame experts yielded significant Elo gains, confirming their effectiveness in capturing the strategic and tactical patterns of their respective phases over the baseline model. In contrast, the opening expert offered little to no improvement and occasionally underperformed relative to the baseline.
This underperformance appears linked to several data-related factors. The opening dataset lacked sufficient diversity, which—combined with a mismatch between training and validation distributions—likely led to overfitting on specific openings. Additionally, the opening expert's limited exposure to downstream consequences during training may have hindered its ability to accurately evaluate early-game positions in context.

These results illustrate that not all game phases benefit equally from specialization and highlight the importance of phase-specific data quality. In modular systems like M2CTS, expert utility depends not only on model capacity but also on having sufficiently diverse and contextually relevant data aligned with the task's structure. This emphasizes the need for targeted data curation and evaluation design when deploying modular learning systems in structured domains like chess. \\

\begin{figure}[tb]
    \centering
    \ifarxiv
    \includegraphics[width=0.9\linewidth]{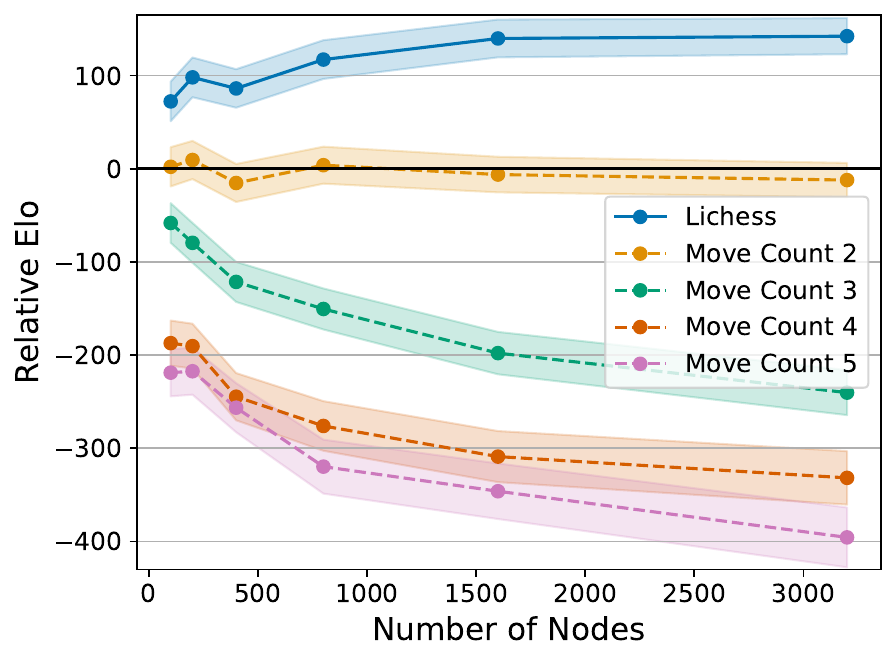}
    \else
    \includegraphics[width=0.9\linewidth]{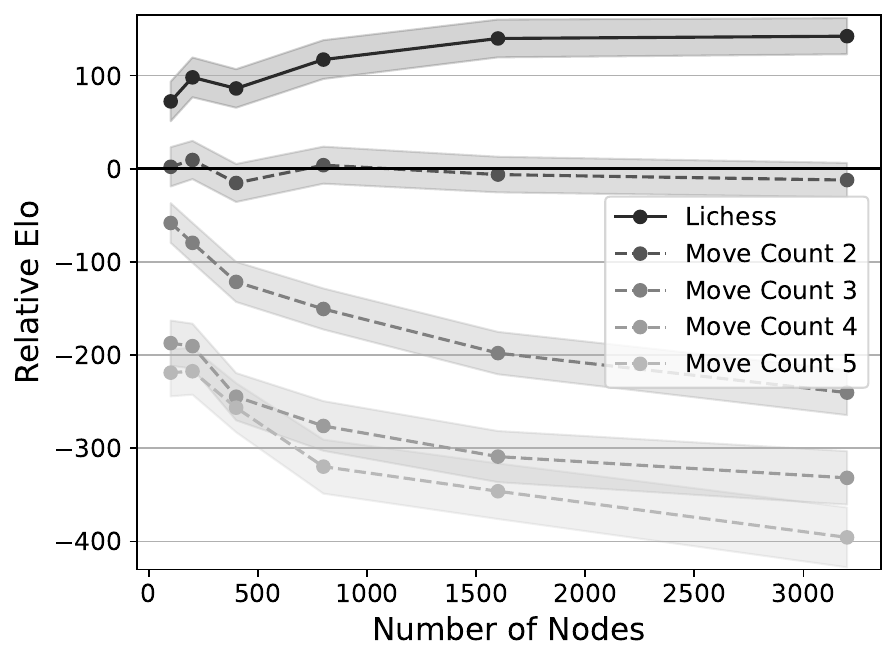}
    \fi
    \caption{{Game phase definitions strongly influence the MoE performance.} Our analysis shows that simplistic approaches, such as relying only on move counts, lead to suboptimal phase categorization and consequently to inferior performance. In contrast, adopting the Lichess phase definition leads to a significant performance improvement.}
    \label{fig:phasedefinition_experiment}
\end{figure}

\textbf{Effective gating requires semantically meaningful phase definitions informed by domain understanding.}
We began with the traditional three-phase division of chess—opening, middlegame, and endgame—based on long-established concepts from chess theory and practice. This classical structure reflects intuitive shifts in strategy and board dynamics recognized by both human players and chess literature. However, we questioned whether such domain-specific knowledge is necessary for defining effective phases in our framework. Could phase definitions be simplified, or even derived without expert insight?

To explore this, we experimented with alternative partitions based purely on the move number, aiming to test if temporal cues alone could serve as a proxy for strategic transitions. Specifically, we divided the dataset into 2, 3, 4, and 5 equal-sized segments according to move count, creating multiple versions of phase definitions with similar data volume per partition. Figure~\ref{fig:phasedefinition_experiment} presents the Elo comparisons across these configurations.

The results show that increasing the number of phases does not consistently improve performance; instead, alignment between phase boundaries and meaningful strategic shifts in the game is essential. Notably, the 3-phase model based on Lichess metadata outperforms a 3-phase model using only the move counter, indicating that domain-informed segmentation is more effective than naive time-based clustering.

These findings suggest that building an effective gating mechanism requires more than just uniform splits—it benefits from conceptually grounded phase definitions that reflect actual shifts in strategic intent. This insight extends beyond chess, reinforcing a broader principle: when designing modular systems, the quality of task segmentation plays a critical role in downstream performance.\\

\begin{table*}[tb]
\centering
\begin{tabular}{ll|cccc}
\toprule
                                &   & \multicolumn{4}{c}{\textbf{Testset Loss}}                                                                                                                                                                                                                                                                                    \\  
\multirow{-2}{*}{\textbf{Approach}} & \multirow{-2}{*}{\textbf{Expert Model}}                     & {Opening}                     & {Middlegame}                    & {Endgame} & {No-Phases}  \\        

\midrule
\textbf{MCTS} & - & {0.9269} & {1.5043} & {1.3523} & {{1.2897}} \\
\midrule
\textbf{M2CTS} & Opening & \textbf{0.9169} & 2.4030 & 3.2433 & 2.2587 \\
& Middlegame & 1.6064 & \textbf{1.4371} & 1.5494 & 1.5260 \\
& Endgame & 2.0097 & 1.6605 & \textbf{1.2594} & 1.6293 \\
& Mixture & \textbf{0.9169} & \textbf{1.4371} & \textbf{1.2594} & {\textbf{1.2245}} \\
\bottomrule
\end{tabular}
\caption{{M2CTS reduces overall loss in chess using game phase specific experts.} This table provides a comprehensive overview of the loss values for each expert's final model across different approaches. Performance is evaluated over four different test sets: opening, middlegame, endgame, and no-phases. In addition to the individual expert losses, we include the loss for the resulting mixture model that includes all three experts. This mixture model determines the appropriate expert based on the phase of the current position in the test set. The most effective model within each learning process for each test set is highlighted in bold.}
\label{tab:short_loss}
\end{table*}

\begin{table*}[tb]
\centering
\begin{tabular}{cc|rr}\\\toprule
\multicolumn{2}{c|}{\textbf{Evaluations per Turn (Nodes/Time)}} & \textbf{CA w. M2CTS} & \textbf{CA w. MCTS} \\
ClassicAra 1.0.4 (CA) & Stockfish 16.1 & \multicolumn{2}{c}{vs. Stockfish 16.1} \\\midrule
3200 nodes & 3200 nodes & $90\pm19.62$ $\bullet$ & $-35\pm18.46$\\  
3200 nodes & 4000 nodes & $19\pm18.40$ $\bullet$ & $-76\pm18.85$\\  
3200 nodes & 5000 nodes & $-39\pm18.30$ $\bullet$ & $-129\pm19.40$\\  
3200 nodes & 6400 nodes & $-108\pm18.41$ $\bullet$ & $-214\pm20.67$\\  
3200 nodes& 12800 nodes & $-253\pm21.65$ $\bullet$ & $-356\pm26.65$\\\midrule  
500 ms & 100 ms & $-144\pm16.50$ $\bullet$ & $-281\pm21.00$\\  
1000 ms & 100 ms & $-95\pm15.60$ $\bullet$ & $-175\pm18.00$\\  
\bottomrule
\end{tabular}
\caption{{M2CTS performs better ($\bullet$) than MCTS against Stockfish.} This table illustrates the outcomes of our M2CTS framework versus the baseline MCTS in an indirect competition of 1000 games, emphasizing M2CTS's significant improvement. Despite not yet surpassing Stockfish, our results demonstrate M2CTS's consistent superiority over the baseline approach.}
\label{tab:stockfish}
\end{table*}

\begin{figure}[tb]
    \centering
    \ifarxiv
      \includegraphics[width=0.9\linewidth]{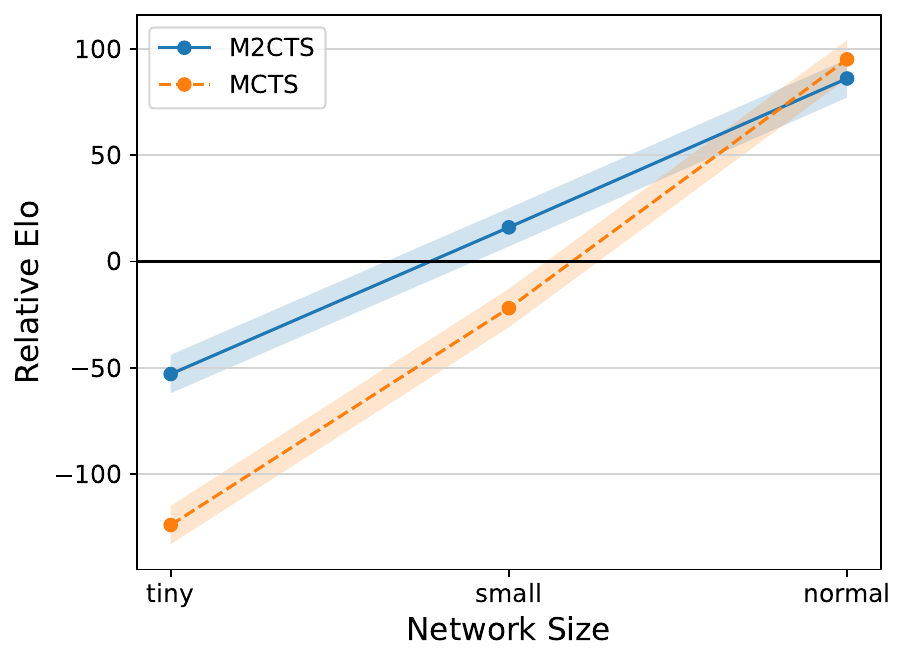}
      \else
      \includegraphics[width=0.9\linewidth]{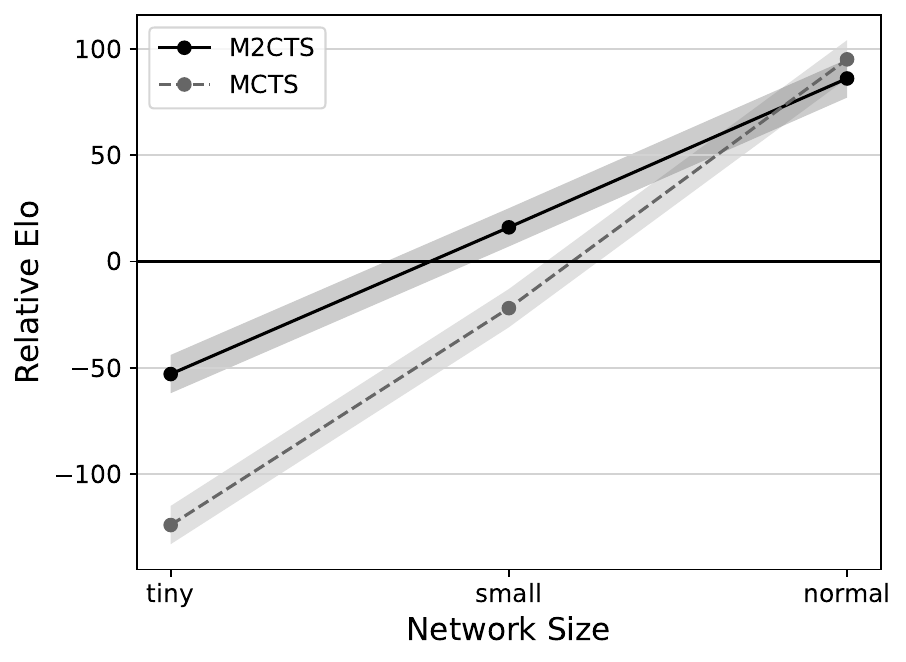}
      \fi
    \caption{{M2CTS using sperated learning outperforms standard MCTS at smaller model sizes.} Elo was measured in a round-robin tournament with a fixed 800 number of nodes per move using different sizes of the AlphaVile~\cite{czech2024representation} architecture. All agents were trained on the KingBase Lite 2019 dataset. 
    }
      \label{fig:scalability}
\end{figure}

\begin{table*}[tb]
    \centering
    \begin{tabular}{l|ccc}
    \toprule
\textbf{Elo Difference between MoE and MCTS}  & \multicolumn{3}{c}{\textbf{Network}} \\  
Dataset & RISEv3.3 & AlphaVile-Tiny &  AlphaVile-Small\\        
\midrule
KingBase Lite 2019 (full)& 117.2 $\pm$ 20.8 $\bullet$ & 35.9 $\pm$ 19.7 $\bullet$&  45.1 $\pm$ 19.3 $\bullet$\\
KingBase Lite 2019 (half-sized)& -109.5 $\pm$ 20.8\hspace{2ex} & -144.7 $\pm$ 21.5\hspace{3.5ex} & -139.0 $\pm$ 21.1\hspace{3.5ex} \\
\bottomrule
\end{tabular}
    \caption{{M2CTS with seperated learning improves with larger dataset size.} Models were trained both on the full KingBase Lite 2019 dataset and a half-sized KingBase Lite 2019 subset. Next, the M2CTS version of each model played against the standard MCTS version using 800 nodes per move. M2CTS performs better ($\bullet$) than standard MCTS when the models were trained on the larger dataset.}
    \label{tab:scalability}
\end{table*}

\textbf{M2CTS with seperated learning requires sufficient data to outperform MCTS in separated learning.}
To understand how dataset size and model capacity affect \ours{}, we conducted two scalability experiments. First, we used the AlphaVile architecture~\cite{czech2024representation}, which is available in multiple sizes (tiny, small, normal), and trained both single-model and MoE variants on our standard dataset. We then measured relative Elo differences between the monolithic and modular approaches (see \cref{fig:scalability}).
In the second experiment, we reduced the dataset to half its original size and repeated training for all network variants. The results (\cref{tab:scalability}) show that while larger models benefit from MoE with sufficient data, their performance deteriorates rapidly when data is limited. In contrast, single-network baselines are more robust under constrained conditions.

These results suggest MoE strategies like separated learning are effective only when each expert has access to enough data. With smaller datasets, the per-expert data becomes too sparse to support reliable learning. As expected, larger networks also require proportionally more data to reach their full potential.

\begin{figure}[tb]
    \centering
    \ifarxiv
    \includegraphics[width=0.9\linewidth]{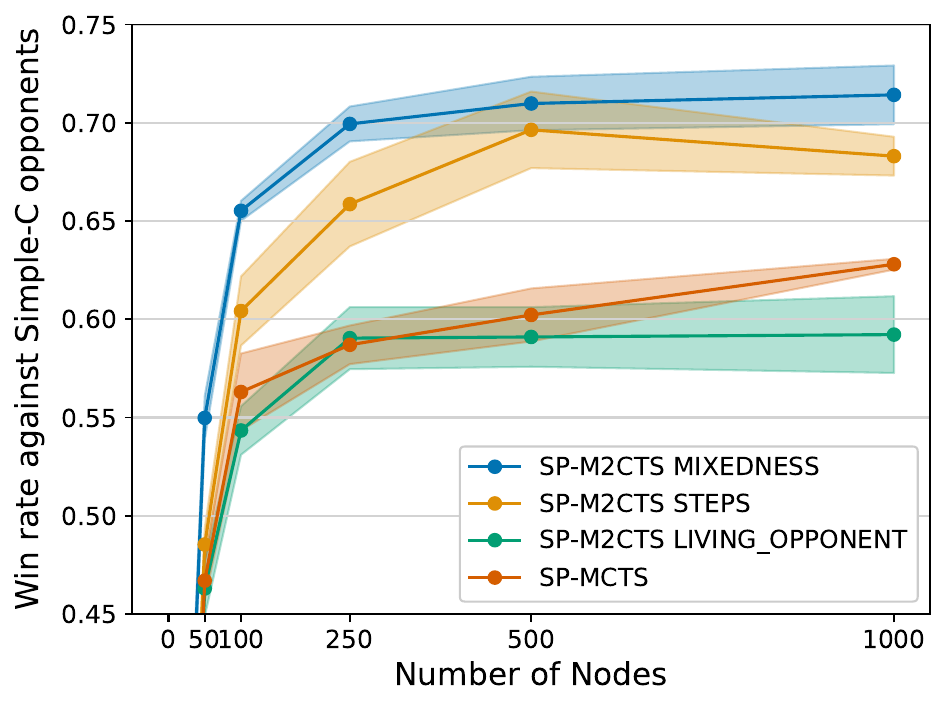}
    \else
    \includegraphics[width=0.9\linewidth]{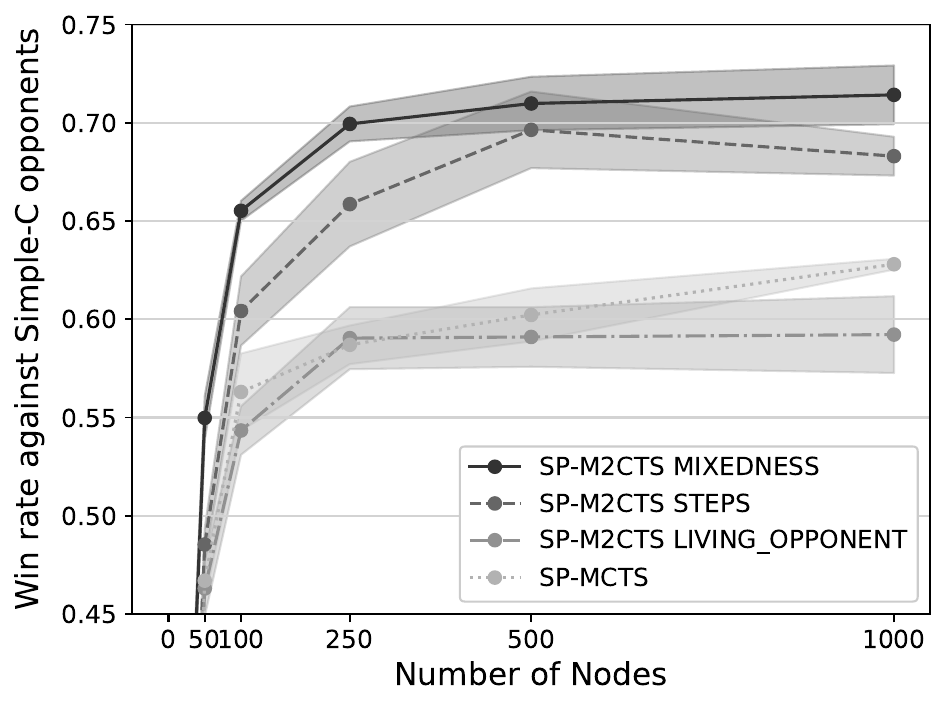}
    \fi
    \caption{{M2CTS improves performance in the Pommerman~\cite{resnick2018pommerman} environment for small networks.} All networks were trained for five epochs using separated learning on a dataset generated by a simple agent heuristic. The win rate is measured in the ``free-for-all'' game mode, where our agent plays three simple agents using single player (SP) M2CTS/MCTS search.}
    \label{fig:moe_pommerman}
\end{figure}

\subsection{Beyond Chess (Q3).}
\begin{figure}[tb]
    \centering
    \ifarxiv
    \includegraphics[width=0.65\linewidth]{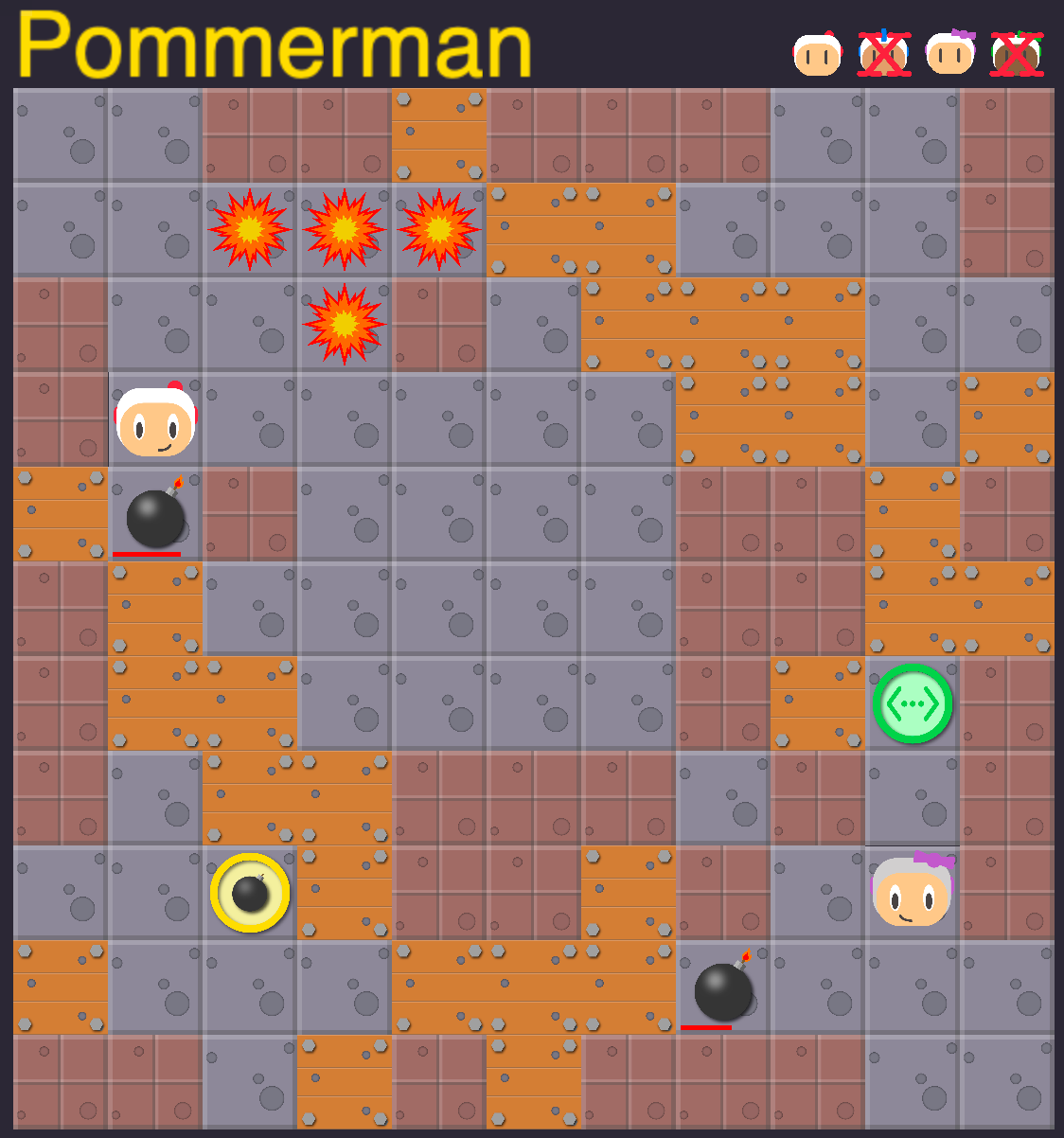}
    \else
    \includegraphics[width=0.65\linewidth]{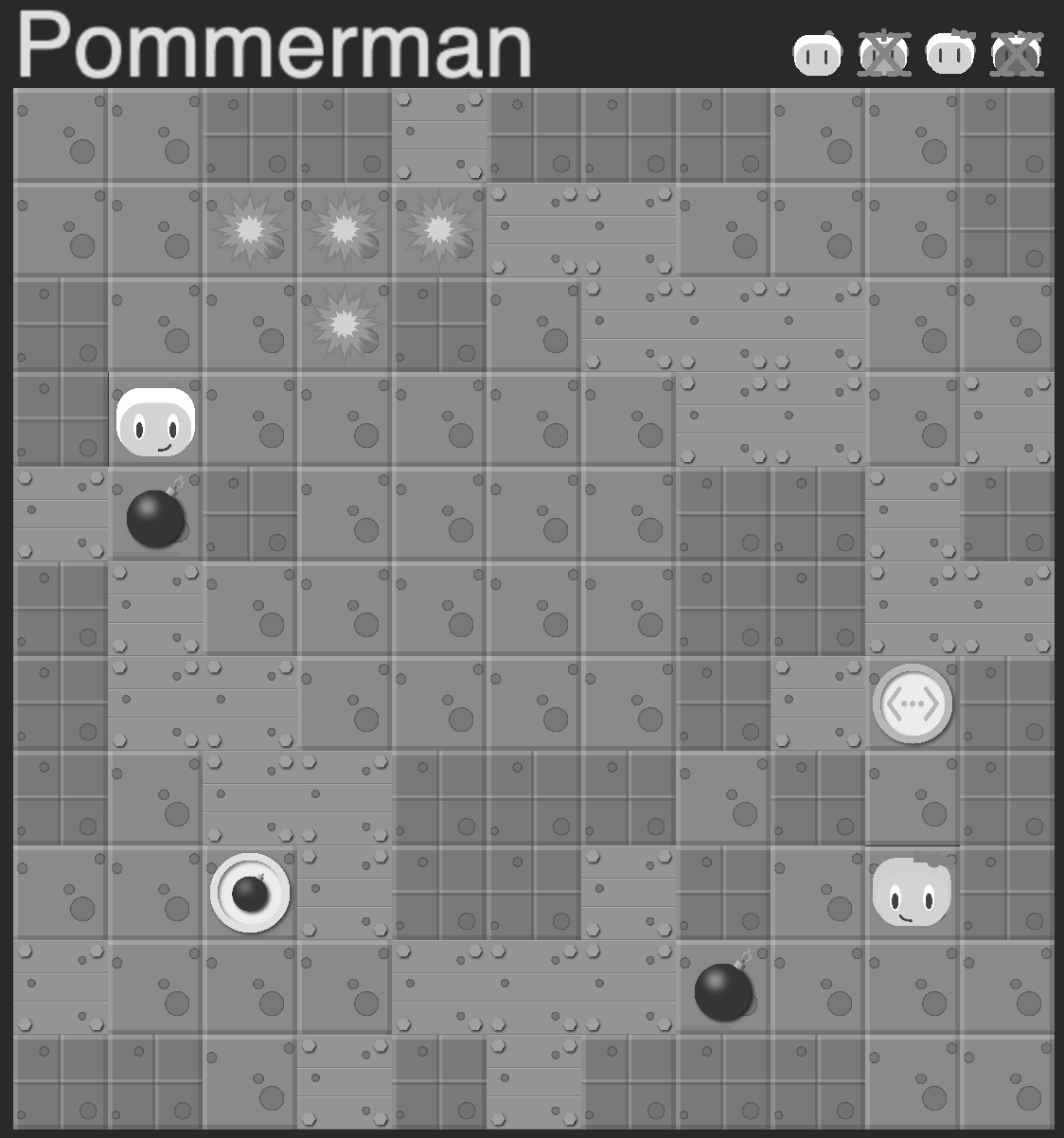}
    \fi
    \caption{Exemplary game position of the Pommerman~\cite{resnick2018pommerman} environment. The goal is to defeat other players by placing bombs, destroying boxes and collecting power-ups.}
    \label{fig:pommerman}
\end{figure}

\textbf{M2CTS can outperform MCTS in Pommerman.}
To test whether the benefits of modular, phase-aware decision-making generalize beyond turn-based board games like chess, we evaluate \ours{} in the Pommerman environment~\cite{resnick2018pommerman}. Pommerman is a real-time, multi-agent environment inspired by the classic game Bomberman. It features partially observable, dynamic interactions in an 11$\times$11 grid world where agents must place bombs to destroy boxes, collect power-ups, and eliminate opponents. This setting provides a useful testbed for studying search based algorithms under complex multiplayer conditions. An example game state is shown in \cref{fig:pommerman}.
%
We trained small-scale RISEv3.3 networks (one base channel, four expanded channels, two residual blocks) using separated learning for five epochs. Training data was generated via self-play among simple heuristic-based agents, simulating a low-resource but diverse gameplay scenario. The purpose of this setup was to test whether phase-aware specialization could still offer a benefit when data and model capacity are highly constrained.

To adapt M2CTS to this setting, we defined three different heuristics which we used for game-phase separation in Pommerman. 

\begin{enumerate}
[leftmargin=20pt,itemsep=2pt,parsep=2pt,topsep=1pt,partopsep=2pt] 
    \item \textbf{\textsc{Steps}}: Divides the game based on time progression (early, mid, late based on number of actions taken).
    \item \textbf{\textsc{Mixedness}}: Measures the manhatten distance to the next opponent and seperates the game phases depending on their proximity.
    \item \textbf{\textsc{Living Opponent Count}}: Segments the game by the number of opponents still alive, reflecting strategic shifts in threat level and exploration focus.\\

\end{enumerate}

Each variant was evaluated in the "free-for-all" (FFA) mode, where our M2CTS agent played against three heuristic-based opponents. As shown in \cref{fig:moe_pommerman}, M2CTS achieved higher win rates in two out of the three configurations, with improvements ranging from 5\% to 11\% over the baseline MCTS agent using a single network. The best-performing variants were \textsc{Steps} and \textsc{Mixedness}, which implicitly captured meaningful strategic transitions. In contrast, the \textsc{Living Opponent Count} heuristic led to unstable phase transitions and inconsistent improvements, likely due to its delayed responsiveness and limited granularity during fast-paced gameplay.

These findings highlight that phase-aware modularity is not confined to structured games like chess, but can extend to complex, multi-agent environments—provided that phase definitions reflect meaningful transitions in game dynamics. Even with small models and limited data, separating learning by coarse but meaningful gameplay structures can yield measurable performance benefits.

\section{Related Work}
Our work brings together advances in MoE models, RL and MCTS, applying them to the structured and well-studied domain of chess. While MoE has gained significant traction in large-scale language models~\citep{ShazeerMMDLHD17, FedusZS22, jiang2024mixtral},
its integration into search-based decision-making frameworks such as AlphaZero~\citep{AlphaZero} remains underexplored. To our knowledge, this work is the first to combine MoE with MCTS in a fully modular, phase-aware framework for strategic games.
In the context of games, MoE has been successfully applied to complex environments like Settlers of Catan~\citep{Dobre2017CombiningAM}, where expert models were trained on heterogeneous datasets to capture different game conditions. In chess, \citet{McIlroy-YoungW022} explored personalized MoE models, fine-tuned to the playing styles of individual human players. These studies highlight the flexibility of MoE in modeling specialized behaviors and adapting to diverse strategic contexts.

Our work is also informed by the long-term evolution of chess engines. The transition from handcrafted heuristics to deep neural networks—exemplified by AlphaZero~\citep{AlphaZero} and NNUE-augmented Stockfish~\citep{nasu2018efficiently}—has yielded strong performance gains, though often at the cost of transparency and alignment with traditional human strategies~\citep{palsson2023unveiling, mcgrath2022acquisition}. Integrating game-phase definitions, a long-standing concept in human chess understanding, has been explored in earlier RL-based engines as well~\citep{block2008using}. 

Our staged learning strategy is inspired by the principles from curriculum learning~\citep{bengio2009curriculum} and task decomposition~\citep{andreas2017modular}, where complex behaviors are learned through progressive specialization and structured modularity.

\section{Discussion}

This work demonstrates that aligning model structure with domain-specific temporal progression—through phase-aware experts and modular search—can improve the performance in strategic environments like chess. The success of separated learning in supervised settings underscores the value of architectural simplicity when data is plentiful and task boundaries are well-defined. Conversely, staged learning offers benefits in reinforcement learning, where early generalization supports more stable exploration.
Our results also highlight that not all forms of decomposition are equally beneficial. Expert effectiveness varies by phase, and naive gating strategies based solely on time steps fall short of domain-informed alternatives. This suggests that modular designs are most effective when they respect underlying task structure, rather than assuming uniform segmentation.
While M2CTS is designed for chess, our Pommerman results suggest broader applicability to multi-agent, partially observable domains—provided that phase transitions reflect meaningful shifts in strategy or game dynamics. The framework offers a path toward integrating learned specialization with planning and suggests that modularity can complement search even under tight capacity and data constraints.

Overall, M2CTS contributes to a growing line of work exploring how modular learning can bring structure and adaptability into decision-making systems—bridging the gap between end-to-end learning and classical notions of phase-aware, human-aligned reasoning.

\subsection*{Limitations}
\label{sec:limitations}

While M2CTS improves performance, it has practical limitations. MoE models can overfit when data is limited, especially as each expert sees only a subset. Our method also increases memory usage since all experts are loaded during search—manageable at our scale but potentially costly with larger models. Though our majority-voting mechanism scales well up to batch sizes of 64, its performance beyond that is untested. Finally, our hand-crafted gating relies on domain-specific phase definitions, limiting generalization to domains without similar structures.

\section{Conclusion and Future Work}

This work introduces M2CTS, a modular framework that integrates Mixture-of-Experts into MCTS, guided by phase-specific expert models. In chess, M2CTS achieves up to +122 Elo over baseline MCTS, with consistent gains in the middlegame and endgame. The framework is robust across batch sizes and generalizes to other domains, such as Pommerman.
While performance in the opening phase was less pronounced, this points to opportunities for improvement in early-game modeling and data representation. M2CTS, based on models trained on human expert data, does not yet surpass state-of-the-art engines like Stockfish, but it provides a compelling step towards modular and phase-aware decision systems.
Future work includes developing a learnable gating network to replace static phase definitions, enabling adaptive expert selection without manual tuning. Additionally, evaluating transfer to other games like Go could further test the framework’s generality. Our results suggest that when strategic substructures can be meaningfully segmented, modular MoE models offer a promising approach to combining search with specialization.

\subsection*{Broader Impact.} This research demonstrates the value of modular architectures in decision-making systems, particularly where domain structure can guide specialization. By combining MoE and MCTS in a principled way, we show how learned specialization can improve performance in complex environments. The ideas explored here may influence the design of efficient AI systems in other strategic domains beyond games.

\section{Ethics Statement}
Throughout the study, ethical guidelines were strictly adhered, ensuring the respectful use of data of any entities involved, like the usage of the KingBase Lite 2019 chess dataset, a collection of chess games played by human players, for the purpose of training and evaluating artificial intelligence models. We acknowledge the potential for biases in the dataset, such as variations in time periods, regions, or player demographics, however were not able to discover any in our research. We provide algorithm and models to make our results reproducible and transparent. We also encourage other researchers to use our framework in their work.

\ifanonym
\section*{Acknowledgements}
We acknowledge the use of ChatGPT and deepl write for reformularting parts of the paper.
\else
\section*{Acknowledgements}
We thank all our colleagues and students for their insights and help with this work. 
We also acknowledge the use of ChatGPT and deepl write for reformularting parts of the paper.
This work was partially funded by the Hessian Ministry of Science and the Arts (HMWK) through the cluster project ``The Third Wave of Artificial Intelligence -- 3AI''.
\fi

\newpage
\bibliographystyle{plainnat}
\bibliography{bib}

\appendix

\clearpage
\end{comment}

\newpage
\section{Data Availability Statement}
\label{app:data}
All data and code, needed to reproduce our supervised learning chess experiments, can be found online.
The KingBase Lite 2019 dataset can be downloaded from \url{{https://archive.org/details/KingBaseLite2019}}. 
To recreate our dataset split, use the code provided by us in {\url{https://anonymous.4open.science/r/CrazyAra-3F10}}, following the instructions in the \textit{ReadMe} file. To test our approach, we provide some models used in this paper anonymized\footnote{\url{https://drive.google.com/drive/folders/1d8CoQBieNqeEomhbYfI_TifZUbCFyS3O?usp=sharing}, accessed 2024-05-22}. These models will be published after acceptance. 

\section{Lichess game phase definitions}
\label{app:lichess}
\label{sec:lichess}
Lichess is an open-source chess server for online play, study, and analysis. State-of-the-art chess engines power their analysis section, which is why players of all skill levels, from beginner to master, use them on a daily basis. Submitting a game for engine analysis results in getting a report of the game development, including a separation of the game in the three typical phases to teach people where they went wrong and how they could improve. The site uses a more sophisticated system to determine game phases by incorporating several transition criteria.\\

\textbf{Endgame definition.} According to the Lichess implementation, a position belongs to the endgame if the total count of major and minor pieces (queens, rooks, bishops, and knights) is less than or equal to 6. Note that while this resembles a material count criterion, it does not use different relative values for the pieces and instead values them all equally as 1. A potential reason for this approach could be that the complexity of a position is not tied to the relative value of its pieces but rather to their total amount.
Furthermore, the pawn count is irrelevant to the decision.\\

\textbf{Middlegame definition.} A position counts towards the middlegame if it does not qualify as an endgame position and if one of the following three criteria is fulfilled.
The number of major and minor pieces is less than or equal to 10, the backrank of at least one player is sparse, or the total mixedness score of the position is bigger than 150. 
Here, backrank sparseness is defined as having less than four total pieces on rank 1 (for white) or 8 (for black), including the king.
The mixedness score describes how close black and white pieces are to each other. It is calculated by going through all two by two squares of the chess board, starting from the square a1, b1, a2, b2 and ending at the square g7, h7, g8, h8. For each of those two by two squares, we count the number of black and white pieces inside it and assign a score based on the result and the square's location. We then sum up all square scores to get the final mixedness score of a position. The exact implementation can be found in the Lichess repository\footnote{\url{https://github.com/lichess-org/scalachess/blob/master/core/src/main/scala/Divider.scala}, accessed 2024-05-22}.\\

\textbf{Opening definition.} All remaining positions are classified as opening positions.\\

Using the previously described phase definitions may lead to transitions to previous phases (e.g., going back to the opening because the mixedness score has increased again). Therefore, to do a strict separation into three sections, Lichess forbids such transitions and only allows transitions to later phases.

\subsection{Resulting Dataset Statistics}
\label{sec:resulting_dataset}

\begin{figure}[b]
\centering
\parbox{6cm}{
\includegraphics[width=6cm]{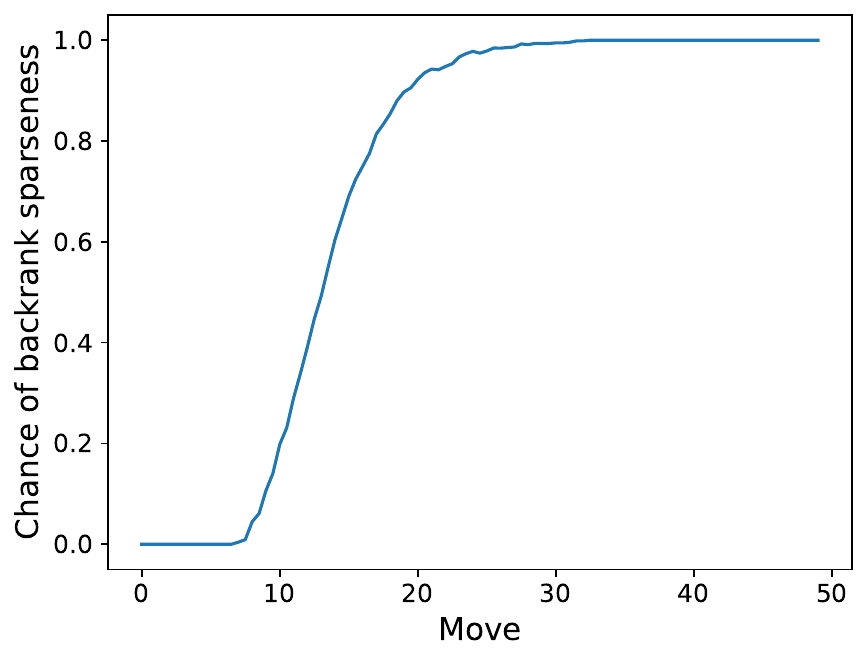}
\caption{The percentage of positions with a sparse backrank by move number. The data is based on all positions of the chess test set.}
\label{fig:results:backrank}}
\qquad
\begin{minipage}{6cm}
    \includegraphics[width=6cm]{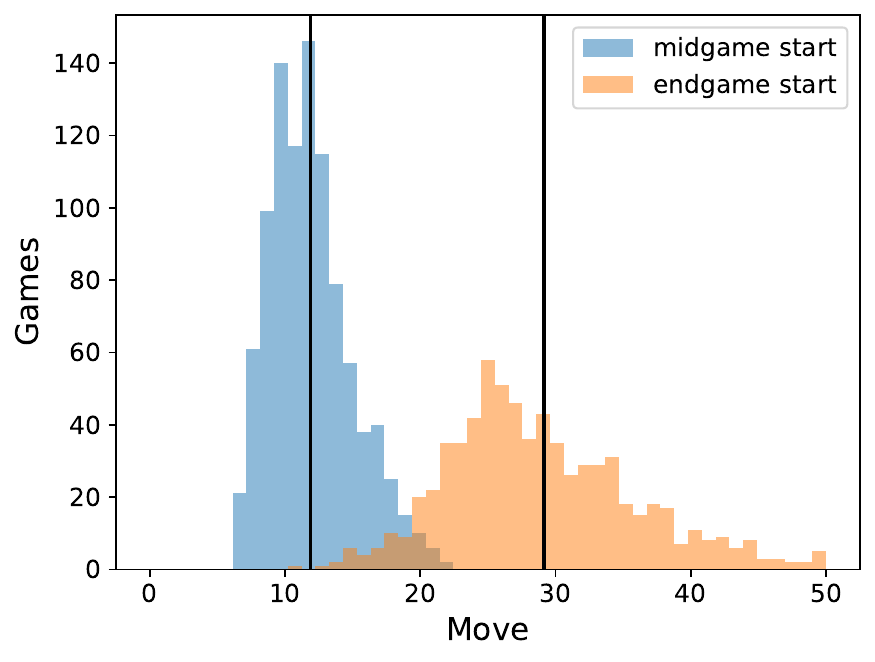}
\caption{The distribution of starting moves of each phase aggregated by move number. The average starting moves of the middlegame (11.89 with an std of 2.966) and the endgame (29.15 with an std of 7.441) are added as vertical lines. The data is based on all games of the chess test set.}
\label{fig:results:phase_start_dist}
\end{minipage}
\end{figure}

After defining the game phases as in Section \ref{sec:lichess}, we examine the distributions of input and output in the resulting datasets for each phase.

\begin{figure}[tbp]
\centering
\includegraphics[width=.9\linewidth]{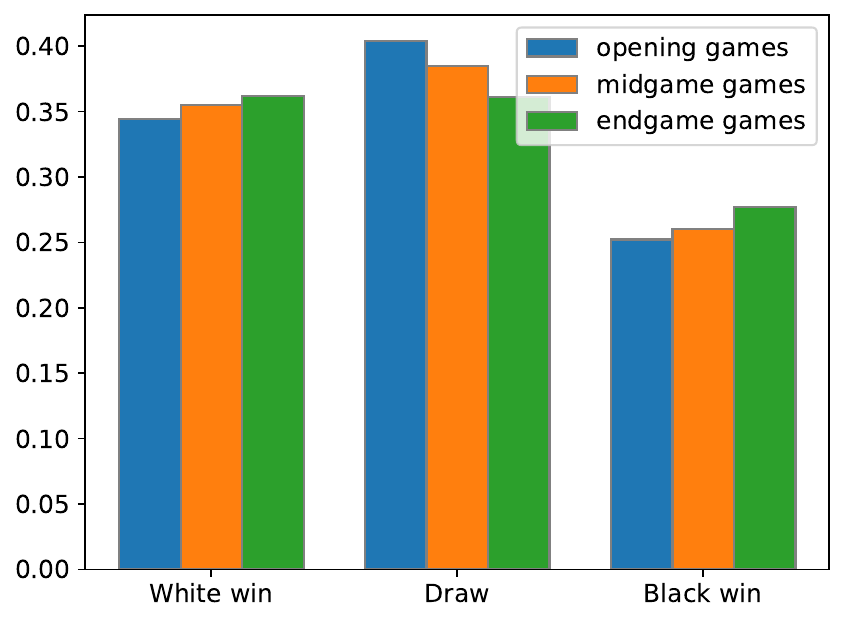}
\caption{Game result distributions in the chess training set, calculated for games reaching each phase.}
\label{fig:results:outcomes_by_phase}
\end{figure}

Figure \ref{fig:results:outcomes_by_phase} shows the distribution of game outcomes in our training set. Every game contributes to the opening phase data, with 34.42\% games won by White, 40.36\% draws, and 25.22\% Black wins. Notably, games extending into the middlegame and endgame phases have a lower draw rate (38.51\% and 36.09\%, respectively) and a higher rate of decisive outcomes.

\begin{figure}
    \centering
    \parbox{6cm}{
    \includegraphics[width=6cm]{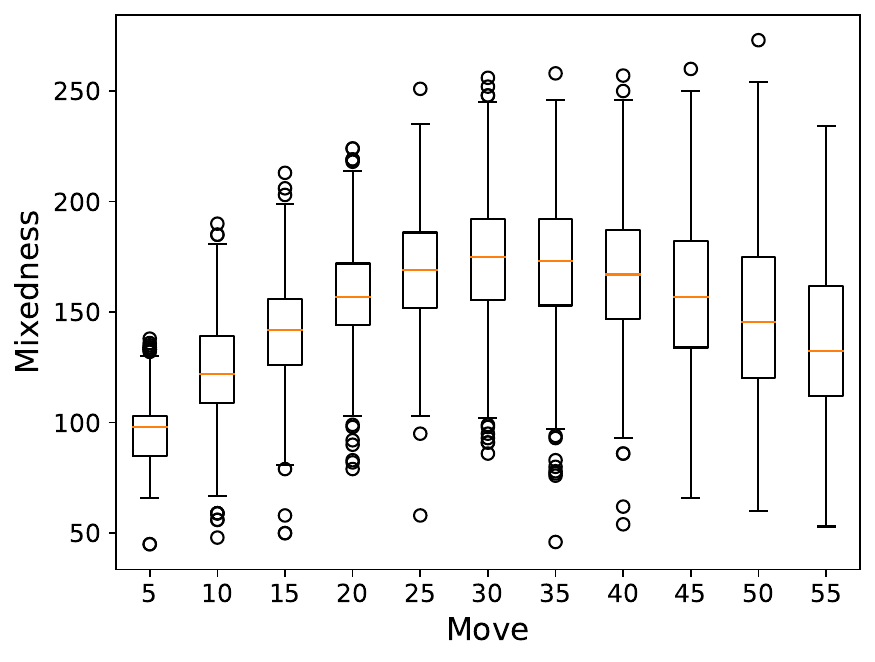}
    \caption{The mixedness scores (according to the definition in Section \ref{sec:lichess}) of chess game aggregated by move number. The data is based on all positions of the chess test set.}
    \label{fig:results:mixedness}
    }
    \qquad
    \begin{minipage}{6cm}
    \includegraphics[width=6cm]{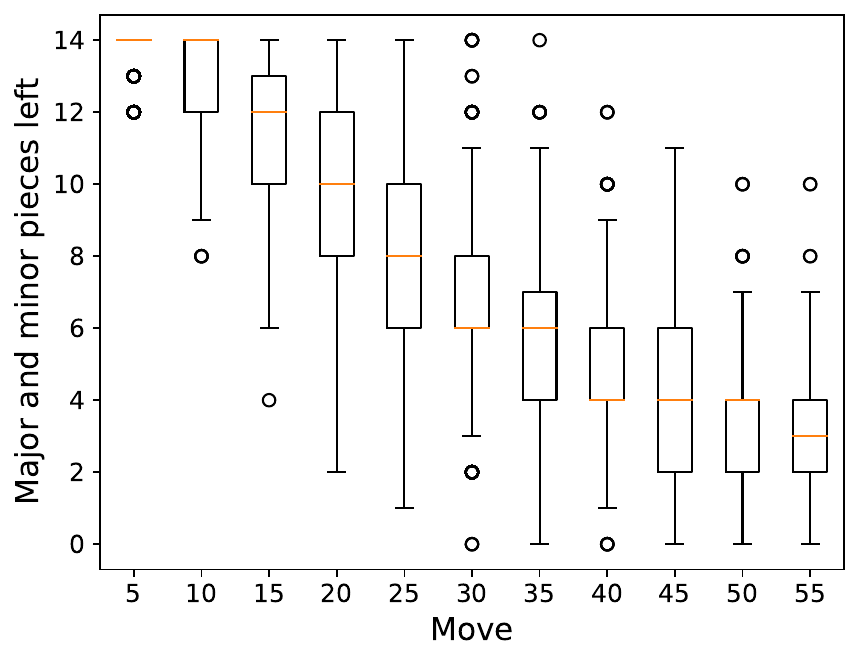}
    \caption{The average number of major and minor pieces left aggregated by move number. The data is based on all positions of the chess test set.}
\label{fig:results:piececount}
    \end{minipage}
\end{figure}

\begin{figure*}[bp]
    \centering
    \subfloat[Games by phase]{\includegraphics[width=.4\textwidth]{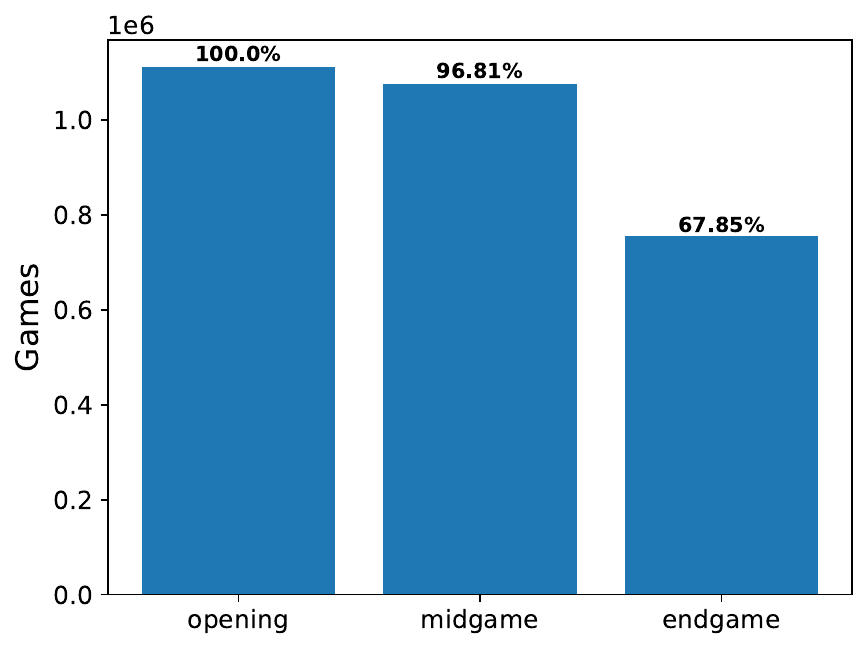}\label{fig:games_by_phase}}
    \subfloat[Positions by phase]{\includegraphics[width=.4\textwidth]{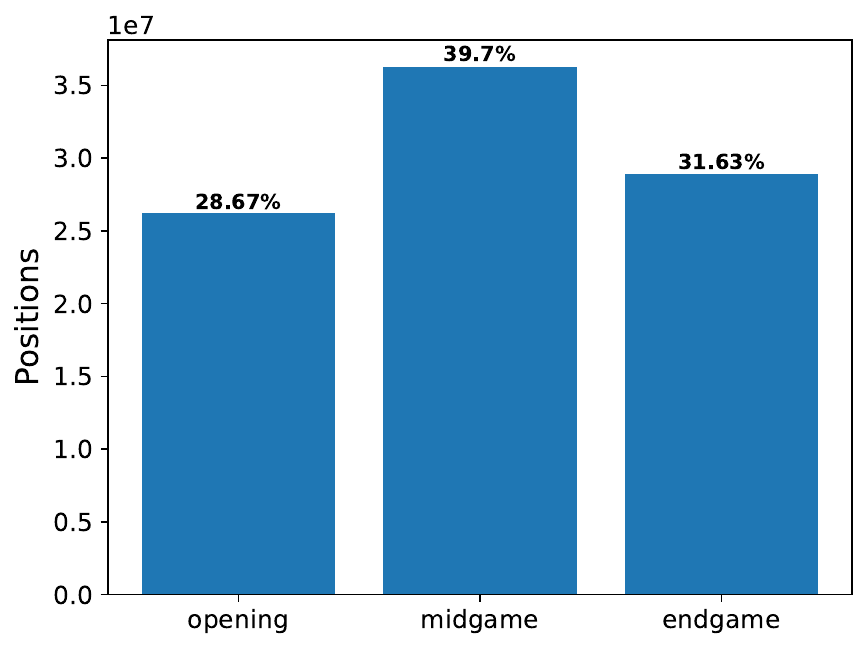}\label{fig:positions_by_phase}}
    \caption{Distribution of games and positions across phases in the chess training set.}
    \label{fig:dataset_balance}
\end{figure*}

Figure~\ref{fig:dataset_balance} illustrates the distribution of game phases. While 96.81\% of the games reach the middlegame, only 67.85\% reach the endgame. Nevertheless, the number of endgame positions (31.63\%) exceeds that of openings (28.67\%), because endgames tend to last longer.

Table~\ref{table:dataset_overview} gives a comprehensive overview of our datasets, the full dataset contains 1,112,647 games and 91,413,951 positions. Table~\ref{table:dataset_overview_games} summarizes the number of games contributing to each phase, highlighting the variance in phase coverage.
\begin{table*}[tbp]
\centering
\begin{tabular}{ccccc}
\toprule
\textbf{Dataset/Phase} & \textbf{No-Phases} & \textbf{Opening} & \textbf{Midgame} & \textbf{Endgame} \\ \midrule
Train Chess            & 91,413,951         & 26,212,273       & 36,287,651       & 28,914,027       \\ 
Val Chess              & 79,042             & 24,566           & 29,333           & 25,143           \\ 
Test Chess             & 85,114             & 24,938           & 31,364           & 28,812           \\ \bottomrule
\end{tabular}
\caption{Overview of dataset sizes by phase, showing the number of positions.}
\label{table:dataset_overview}
\end{table*}

\begin{table*}[tbp]
\centering
\begin{tabular}{cc}
\toprule
\textbf{Dataset/Phase} & \textbf{Months} \\ \midrule
Train Chess            & 2000-01 - 2018-12 (excluding val and test)        \\ 
Val Chess              & 2012-09          \\ 
Test Chess             & 2017-05          \\ \bottomrule
\end{tabular}
\caption{Overview of dataset sizes by games played in which months.}
\label{table:dataset_overview}
\end{table*}

\begin{table*}[tbp]
\centering
\begin{tabular}{ccccc}
\toprule
\textbf{Dataset/Phase} & \textbf{No-Phases} & \textbf{Opening} & \textbf{Midgame} & \textbf{Endgame} \\ \midrule
Train Chess            & 1,112,647          & 1,112,647        & 1,077,136        & 754,899          \\ \bottomrule
\end{tabular}
\caption{Number of games contributing to each phase dataset, indicating phase coverage.}
\label{table:dataset_overview_games}
\end{table*}

\section{Elo Ratings}
\label{app:elo}

The Elo rating system, invented by Arpad Elo, is used to measure relative skill differences between players of a game. Everyone starts with the same arbitrary starting value, and the ratings are adjusted from that point on based on the outcome of finished games. Winning games increases the Elo rating, while losing decreases it. Using the rating of two players $A$ and $B$, it is possible to calculate the expected score of player $A$ (and player $B$ by replacing $R_{\mathrm{B}}-R_{\mathrm{A}}$ with $R_{\mathrm{A}}-R_{\mathrm{B}}$):
\begin{equation}\label{formula:elo_exp}
    E_{\mathrm{A}}=\frac{1}{1+10^{\left(R_{\mathrm{B}}-R_{\mathrm{A}}\right) / 400}}.
\end{equation}

\clearpage
\section{Sample weights for the Weighted Learning Approach}
We carry out experiments for two different a values (\(a \in \{4, 10\}\)) to see whether a big
weight difference (more expert specialization) or a small weight difference (more expert
similarity) leads to better overall performance. The values of the resulting normalized
sample weights can be found in \cref{table:ainWL}.

\begin{table*}[tbh]
\centering
\begin{tabular}{ccccccc}
\toprule
\textbf{a}          & \textbf{Expert Main Phase} & \textbf{$w_{\text{main}}$} & \textbf{$w_{\text{other}}$} & \textbf{$w_{\text{opening}}$} & \textbf{$w_{\text{midgame}}$} & \textbf{$w_{\text{endgame}}$} \\ \midrule
\multirow{3}{*}{4}  & opening                    & \textbf{2}       & 0.5               & \textbf{2}    & 0.5           & 0.5           \\ 
                    & midgame                    & \textbf{2}       & 0.5               & 0.5           & \textbf{2}    & 0.5           \\ 
                    & endgame                    & \textbf{2}       & 0.5               & 0.5           & 0.5           & \textbf{2}    \\ \hline
\multirow{3}{*}{10} & opening                    & \textbf{2.5}     & 0.25              & \textbf{2.5}  & 0.25          & 0.25          \\ 
                    & midgame                    & \textbf{2.5}     & 0.25              & 0.25          & \textbf{2.5}  & 0.25          \\ 
                    & endgame                    & \textbf{2.5}     & 0.25              & 0.25          & 0.25          & \textbf{2.5}  \\ \bottomrule
\end{tabular}
\caption{Sample weights for the Weighted Learning Approach}
\label{table:ainWL}
\end{table*}

\section{Extended Experimental Setup}
\label{app:extendedSetup}
\subsection{Datasets}
The KingBase Lite 2019 dataset is a comprehensive collection of chess games, renowned for featuring high-quality matches played by grandmasters and skilled players. Typically provided in Portable Game Notation (PGN) format, it encompasses detailed information on moves, players, dates, and tournament specifics. Widely employed in computer chess research, machine learning, and artificial intelligence, the dataset is valued for studying advanced chess strategies, training and testing chess engines, and analyzing player performance.

Our chess datasets are built based on the Kingbase Lite 2019 database. This database includes over one million chess games from players rated at least 2200 Elo. We filter out all games shorter than five moves in total as these games are either quickly arranged draws or errors in the database and, therefore, unreliable sources.
In order to create the training and evaluation data for our experiments, we build four datasets. The first dataset, which we will refer to as "no-phases", contains all positions left after filtering the database as described above.
For the remaining three datasets ("opening", "midgame" and "endgame"), we strictly split the database into one part for each of the three phases. This splitting is carried out by going through every game in the dataset and removing all board positions that do not belong to the phase for which we are currently preparing the dataset. This procedure leads to some games being completely excluded for a specific part if they contain no position of the particular phase. For example, about 32 percent of games ended before the endgame began, therefore contributing no position for the endgame dataset.

\subsection{Neural Network Architecture and Input Representation}

\begin{figure}[t]
   \includegraphics[width=0.8\linewidth]{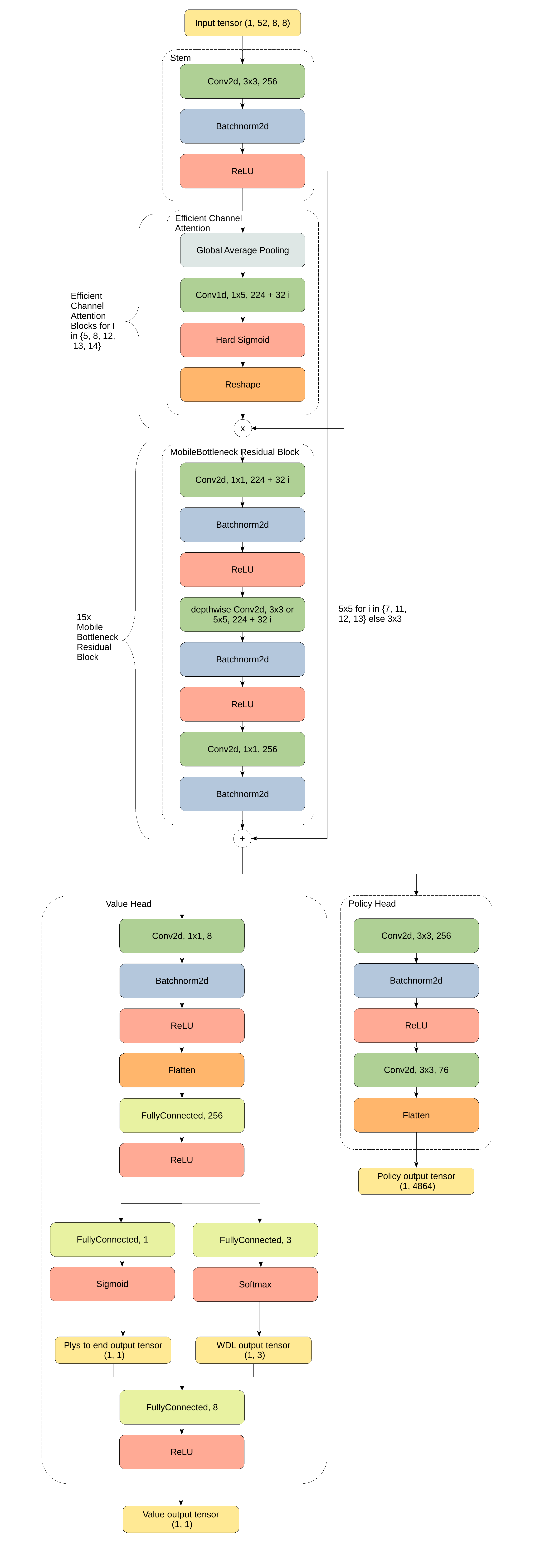}
        \caption{{RISEv3.3 architecture design.} The architecture comprises of 15 Mobile Bottleneck Residual Blocks~\cite{howard2019searching} of which 5 Blocks are preceeded by an Efficient Channel Attention Block~\cite{wang2020eca}. Finally, the output of the model is processed using a value and policy head.}
    \label{fig:rise_v3.3}
\end{figure}

\begin{table*}[tbh]
\centering
\begin{tabular}{lccl}
\toprule
%
\label{tab:input_representation}
\textbf{Feature} & \textbf{Planes} & \textbf{Type}                         & \textbf{Comment}                                                            \\ \midrule
P1 pieces                          & 6                                & bool                      & order: \{\texttt{PAWN}, \texttt{KNIGHT}, \texttt{BISHOP}, \texttt{ROOK}, \texttt{QUEEN}, \texttt{KING}\}                                     \\
P2 pieces                          & 6                                & bool                      & order: \{\texttt{PAWN}, \texttt{KNIGHT}, \texttt{BISHOP}, \texttt{ROOK}, \texttt{QUEEN}, \texttt{KING}\}                                     \\
Repetitions\textsuperscript{*}                      & 2                                & bool                      & \multicolumn{1}{l}{how often the board positions has occurred} \\
En-passant square                 & 1                                & bool                      & the square where en-passant capture is possible                        \\
P1 castling\textsuperscript{*}                      & 2                                & bool                      & binary plane, order: \{\texttt{KING\_SIDE}, \texttt{QUEEN\_SIDE}\}                                      \\
P2 castling\textsuperscript{*}                      & 2                                & bool                      & binary plane, order: \{\texttt{KING\_SIDE}, \texttt{QUEEN\_SIDE}\}                                      \\
No-progress count\textsuperscript{*}                & 1                                & int                      & sets the no progress counter (FEN halfmove clock)                                \\
Last Moves &	16 & bool &	origin and target squares of the last eight moves \\
is960\textsuperscript{*}  &	1 & bool  & if the 960 variant is active \\
{P1 pieces} &	{1} & {bool} &	grouped mask of all P1 pieces \\
{P2 pieces} &	{1} & {bool} &	grouped mask of all P2 pieces \\
{Checkerboard} &	{1} & {bool} &	chess board pattern \\
{P1 Material difference\textsuperscript{*}} &	{5} & {int} &	{order: \{\texttt{PAWN, KNIGHT, BISHOP, ROOK, QUEEN}\}} \\
{Opposite color bishops\textsuperscript{*}} & {1} & {bool} &	{if they are only two bishops of opposite color} \\
{Checkers} &	{1} & {bool} &	{all pieces giving check} \\
{P1 material count\textsuperscript{*}} &	{5} & {int} &	{order: \{\texttt{PAWN, KNIGHT, BISHOP, ROOK, QUEEN}\}} \\ 
\midrule
Total                             & 52                               &                              &                                                                                           \\
\bottomrule
\end{tabular}
\caption{Input representation, taken from \citet{czech2024representation}. Features are encoded as binary maps, and specific features are indicated with $\ast$ as single values applied across the entire $8\times 8$ plane. $*$ represent scalar planes only saving one value over all 64 bits of the plane. The historical context is captured as a trajectory spanning the last eight moves. Overall the input representation consists of 52 planes. }
\label{app:fig:input}
\end{table*}

Since our approach is based on and compared to the CrazyAra approach, by \citet{Czech2020CrazyAra}, we use the same model architecture called RISE in its current version (RISEv3.3). This architecture is in turn an improved ResNet architecture, how they were used in AlphaZero. More information about the architecture can be found in \cref{fig:rise_v3.3}. As loss function we used 
\begin{equation}
    \small \ell = -\alpha (\mathbf{WDL}^{\top}_\text{t} \log \mathbf{WDL}_\text{p}) - \boldsymbol{\pi}^{\top} \log \boldsymbol{p} + \beta (ply_\text{t} - ply_\text{p})^2 + c \|\theta\|_{2}^{2}
\end{equation}
introduced in the work of \citet{czech2024representation}. Within the approach our MCTS implementation follows the work by \citet{KocsisS06, coulom06mcts} with the adaptations, done by \citet{AlphaZero}, e.g., using the PUCT formula instead of UCT. 

Similar to AlphaZero \citep{AlphaZero} or CrazyAra \citep{Czech2020CrazyAra}, we represent the game state in the form of a stack of so-called levels or planes. The complete stack of planes can be found in Table~\ref{app:fig:input} and is taken from \citet{czech2024representation}. Each layer or plane represents a channel describing one of the input features in the current state. Each plane is encoded as a map with $8\times8$ bits. We hereby distinguish between two types of planes, bool and int. In a bool plane each bit is representing a different field of a chessboard, e.g., if a pawn is placed on this field. In some cases, like \textit{repetitions}, a single information is stored in a plane, in these cases all bits of the plane show the same information. In Table \ref{app:fig:input} this is marked with $*$.  Integer or int planes have the same functionality but instead of 0 and 1, they store integer, like how many pawns are left on the board.
The output of our network, also follows \citet{AlphaZero} and is described as the expected utility of the game position, represented by a numeric value in the range of $[-1,1]$, often called value, and a distribution over all possible actions, called policy $\boldsymbol{\pi}$. 

\subsection{Reproducibility and Hyperparameters}
\label{app:hyperparameter}

\begin{table*}[h!]
\centering
\begin{tabular}{llll}
\toprule
\textbf{Hyperparameter} & \textbf{Value} & \textbf{Hyperparameter} & \textbf{Value}  \\
\midrule
max learning rate       & 0.14           & value loss factor       & 0.01           \\
min learning rate       & 0.00001        & policy loss factor      & 0.988          \\
batch size              & 2048           & wdl loss factor         & 0.01           \\
max momentum            & 0.95           & plys to end loss factor & 0.002          \\
min momentum            & 0.8            & stochastic depth probability & 0.05      \\
epochs                  & 7              & pytorch version         &  1.12.0    \\
optimizer               & NAG           & spike thresh            & 1.5\\
weight decay (wd)       & 0.0001        & dropout rate            & 0.0\\
seed                    & 9             & sparse policy label & True \\

\bottomrule
\end{tabular}
\caption{Hyperparameter configuration for experimental settings using supervised learning. This table provides a comprehensive overview of the essential hyperparameters utilized in our experimental design.}
\label{tab:hyperparams}
\end{table*}

\begin{table*}[h!]
\centering
\begin{tabular}{llll}
\toprule
\textbf{Hyperparameter} & \textbf{Value} & \textbf{Hyperparameter} & \textbf{Value}  \\
\midrule
max learning rate       & \textbf{0.05}           & value loss factor       & \textbf{0.499}           \\
min learning rate       & \textbf{0.000005}        & policy loss factor      & \textbf{0.499}          \\
batch size              & \textbf{512}           & wdl loss factor         & \textbf{0.499}           \\
max momentum            & 0.95           & plys to end loss factor & 0.002          \\
min momentum            & 0.8            & stochastic depth probability & 0.05      \\
epochs                  & 1              & pytorch version         &  1.12.0    \\
optimizer               & NAG           & spike thresh            & 1.5\\
weight decay (wd)       & 0.0001        & dropout rate            & 0.0\\
seed                    & \textbf{1,2,3}             & sparse policy label & \textbf{False} \\

\bottomrule
\end{tabular}
\caption{Hyperparameter configuration for the reinforcement learning setup. Changes compared to the supervised learning setup are marked in \textbf{bold}.}
\label{tab:hyperparams_rl_training}
\end{table*}

\begin{table*}[h!]
\centering
\begin{tabular}{llll}
\toprule
\textbf{Hyperparameter} & \textbf{Value} & \textbf{Hyperparameter} & \textbf{Value}  \\
\midrule
batch size       & 8           & dirichlet $\alpha$       & 0.3           \\
temperature       & 0.8           & dirichlet $\epsilon$       & 0.25           \\
temperature moves      & 15           & nodes      & 800           \\
reuse search tree      & False           & number neural network updates      & 10           \\
new training samples before update & 819200 & replay memory fraction for selection & 0.05  \\
number samples from replay memory for update & 4096000 & & \\
\bottomrule
\end{tabular}
\caption{Hyperparameter generation settings for the reinforcement learning setup.}
\label{tab:hyperparams_rl_training}
\end{table*}

\begin{figure*}[tbp]
    \centering
    \subfloat[Learning rate schedule]{\includegraphics[width=.45\textwidth]{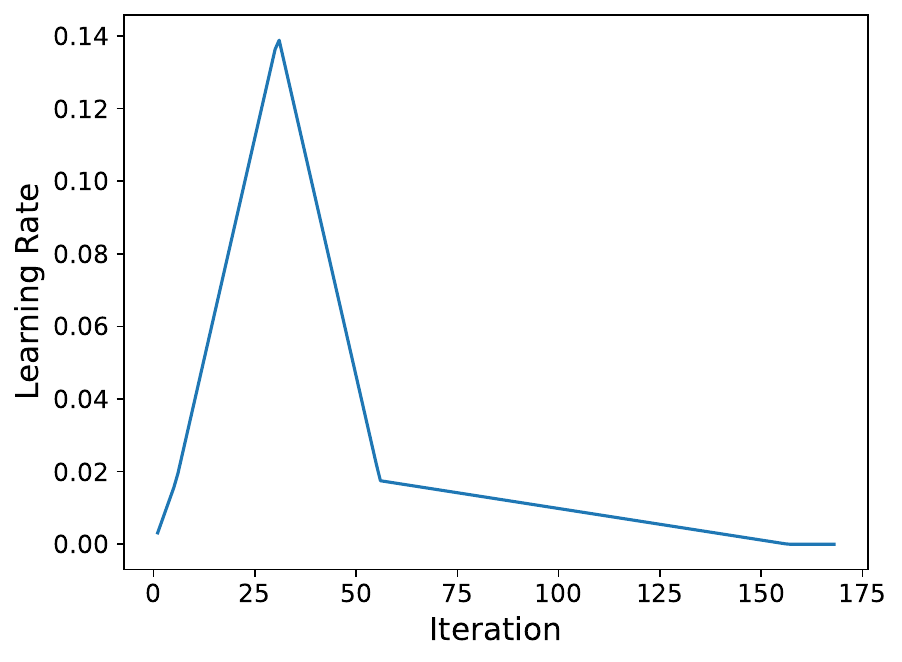}\label{fig:lr_schedule}}
    \subfloat[Momentum schedule]{\includegraphics[width=.45\textwidth]{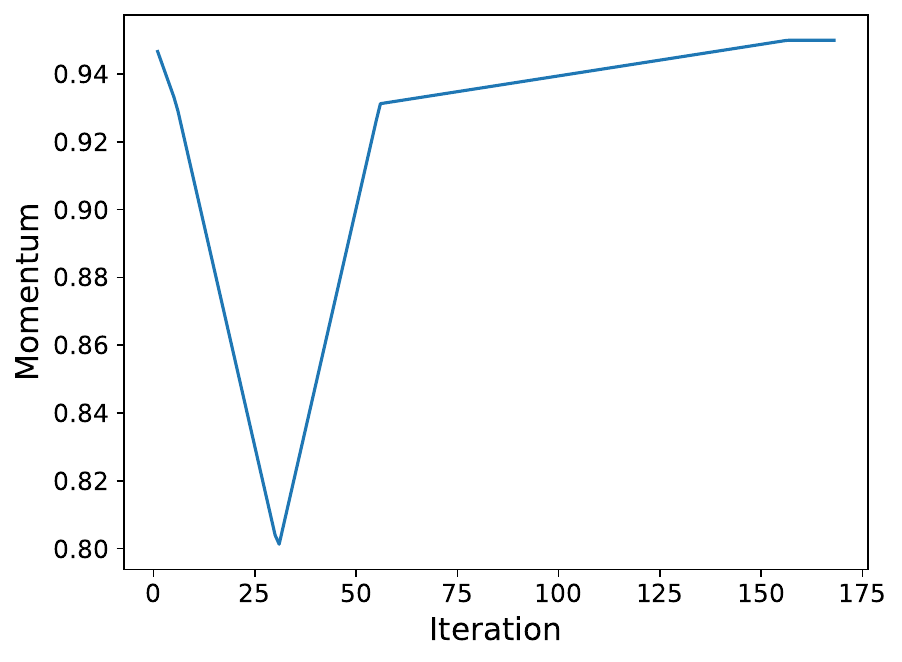}\label{fig:momentum_schedule}}
    \caption{Learning rate and momentum schedules over the course of a training run of a single expert model.}
    \label{fig:schedules}
\end{figure*}
\begin{table*}[tbh!]
    \centering
    \begin{tabular}{lr} \toprule
        \textbf{Hardware/Software} & \textbf{Description}  \\ \midrule
        GPU & 8 $\times$ NVIDIA® Tesla V100 \\
        NGC Container & pytorch:22.05 \\
        GPU-Driver & CUDA 11.4 \\
        CPU & Dual Intel Xeon Platinum 8168 \\
        Operating System & Ubuntu 20.04 LTS \\
        CrazyAra & Release 1.0.4 \\ 
        Backend & TensorRT-8.4.1 \\
        \bottomrule
    \end{tabular}
    \caption{Hard- and software configuration for our experimental section. }
    \label{tab:hardware}
\end{table*}
The list of hyperparameters can be seen in \cref{tab:hyperparams} and the schedules for learning rate and momentum are depicted in \cref{fig:schedules}. An exemplary command for running cutechess\footnote{For detailed information about the cutchess-cli \url{https://manpages.ubuntu.com/manpages/xenial/en/man6/cutechess-cli.6.html}, accessed 2024-05-22} matches with all used parameters can be found in Figure \ref{verb:hyperparams_cutechess}. In order to provide a diversified playing ground, we make use of an opening suite\footnote{\url{https://raw.githubusercontent.com/ianfab/books/master/chess.epd}, accessed 2024-05-22} featuring positions appearing after the first several moves have already been played. These positions are intentionally chosen in a way that each position is imbalanced, with a slight edge for either side to reduce the amount of resulting draws. For each match, we randomly sample 500 opening positions from this opening suite and let our agents play against each other starting from there. In order to take the imbalances of the positions into account, we use each starting position twice so that each agent is playing once as white and once as black. Due to the high computational costs of training, we do not provide any hyperparameter optimization and instead took the proposed settings by \citet{Czech2020CrazyAra}.
\begin{figure*}[tbh!]
\begin{verbatim}
./cutechess-cli -variant standard -openings file=chess.epd format=epd order=random
-pgnout /data/cutechess_res.pgn -resign movecount=5 score=600 -draw movenumber=30
movecount=4 score=20 -concurrency 1
-engine name=ClassicAra_correct_phases cmd=./ClassicAradir=~/CrazyAra/engine/build
option.Model_Directory=/data/model/ClassicAra/chess/correct_phases proto=uci
-engine name=ClassicAra_no_phases cmd=./ClassicAra dir=~/CrazyAra/engine/build
option.Model_Directory=/data/model/ClassicAra/chess/no_phases proto=uci
-each option.First_Device_ID=0 option.Batch_Size=64 option.Fixed_Movetime=0
tc=0/6000+0.1 option.Nodes=100 option.Simulations=200 option.Search_Type=mcts
-games 2 -rounds 500 -repeat
\end{verbatim}
\centering
\caption{Exemplary command for running a cutechess match between two approaches. The \textit{Batch\_Size} is set to 1, 8, 16, 32, and 64. The \textit{Nodes} parameter (values of 0, 100, 200, 400, 800, 1600, 3200) limits the number of nodes that the MCTS can visit per move during its search. The \textit{Simulations} parameter describes the number of simulations per move and is set to double the \textit{Nodes} value. When the \textit{Nodes} parameter is 0, we instead limit the extent of the search by a fixed move time in milliseconds (\textit{Fixed\_Movetime} values of 100, 200, 400, 800, 1600). }
\label{verb:hyperparams_cutechess}
\end{figure*}
Ensuring reproducibility is paramount in scientific research, as it establishes the foundation for the reliability and credibility of study findings. For this, we provide an anonymous version of our code, with a short manual how to run the code in the \textit{ReadMe} as well as our trained models used in the experimental section. The experiments were run on a setup, described in \cref{tab:hardware}, using the NVIDIA GPU Cloud (NGC) docker container for pytorch\footnote{\url{https://catalog.ngc.nvidia.com/orgs/nvidia/containers/pytorch}, accessed 2024-05-22}. As stated before, all needed data is openly available. 
\clearpage

\clearpage
\section{Additional Results}
In this section, we present supplementary findings and extended analyses that complement and enrich the core results discussed in the paper, providing a more comprehensive understanding of the investigated approach. 

\begin{figure*}[b]
    \centering
    \subfloat[Loss]{\centering
      \includegraphics[width=.48\textwidth,trim=0 0 0 0,clip]{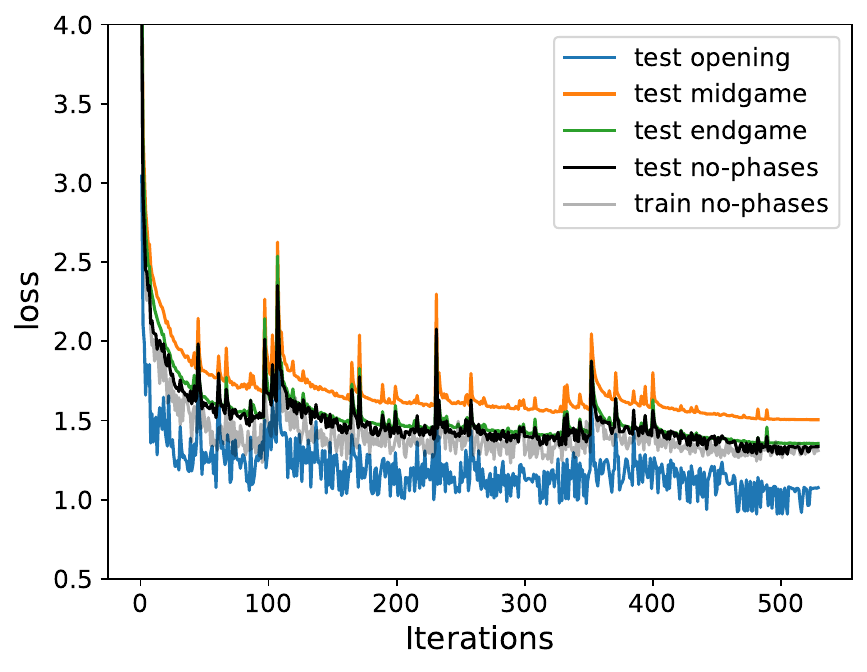}
      \label{fig:correct_no-phases_loss}
    }
    \subfloat[Policy Accuracy]{\centering
      \includegraphics[width=.48\textwidth,trim=0 0 0 0,clip]{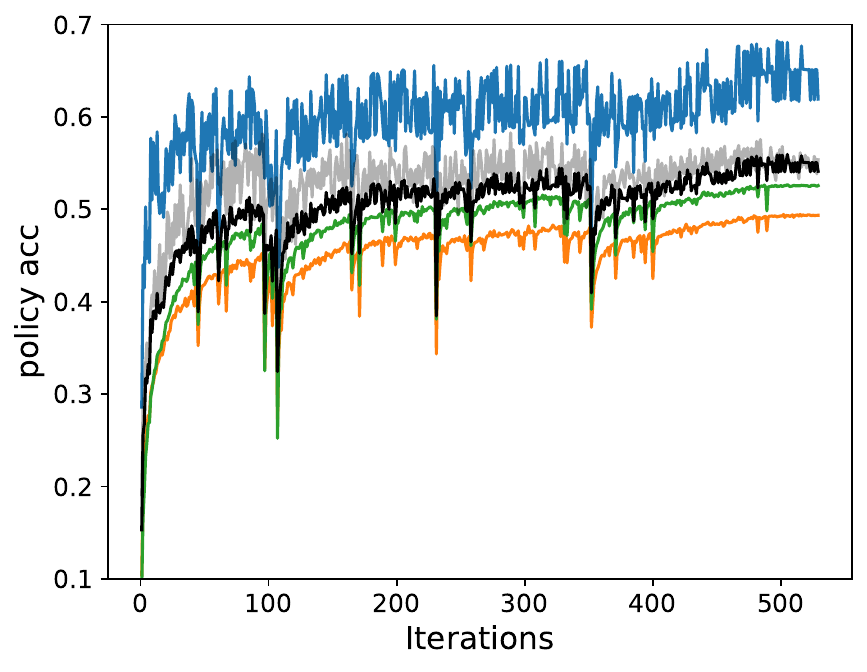}
      \label{fig:correct_no-phases_policy}
    }
    \caption{The loss (Fig. \ref{fig:correct_no-phases_loss}) and policy accuracy (Fig. \ref{fig:correct_no-phases_policy}) over the course of the training process for the \textit{Regular Learning} approach. The metrics were evaluated on the opening (blue), midgame (orange), endgame (green) and no-phases (black) test set. The evaluation on the train set is shown in the background in gray. The values on the x-axis represent iterations, where one iteration is defined as doing backpropagation on one batch of training data.}
    \label{fig:sep_metrics_no-phases}
\end{figure*}

\begin{figure*}[tbh]
    \centering
    \subfloat[Loss ($a=4$)]{\centering
      \includegraphics[width=.48\textwidth,trim=0 0 0 0,clip]{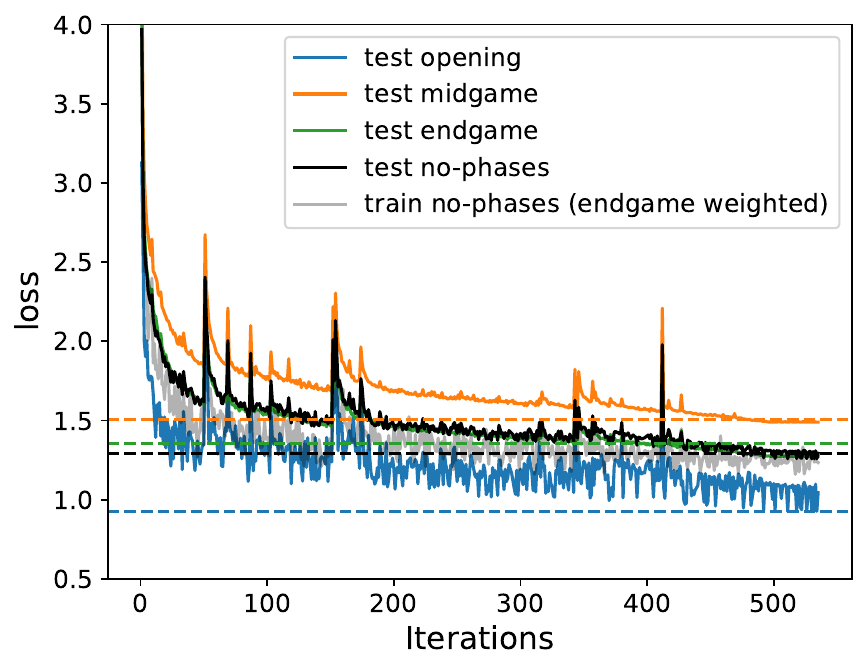}
      \label{fig:weighted025_endgame_loss}
    }
    \subfloat[Policy Accuracy ($a=4$)]{\centering
      \includegraphics[width=.48\textwidth,trim=0 0 0 0,clip]{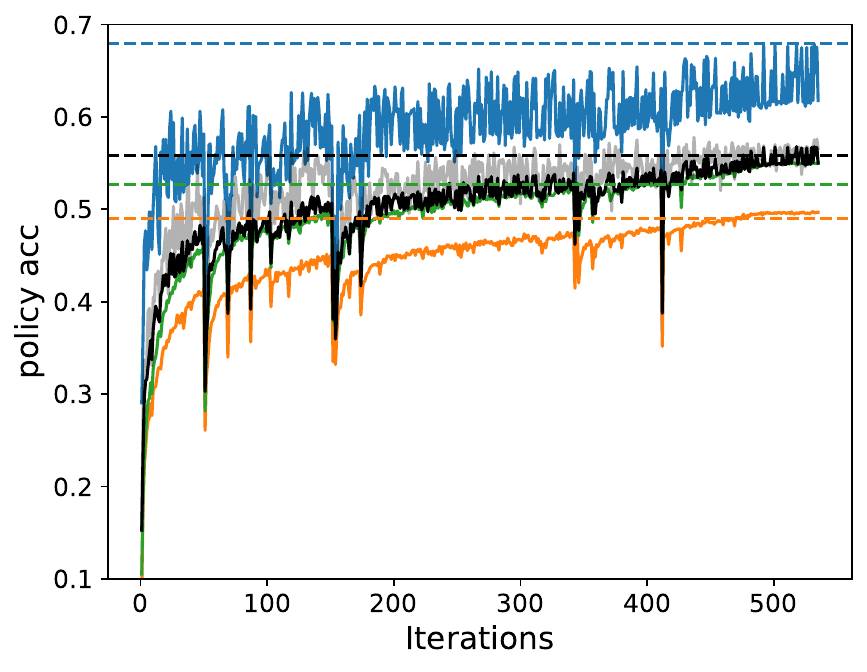}
      \label{fig:weighted025_endgame_policy}
    }\\
    \subfloat[Loss ($a=10$)]{\centering
      \includegraphics[width=.48\textwidth,trim=0 0 0 0,clip]{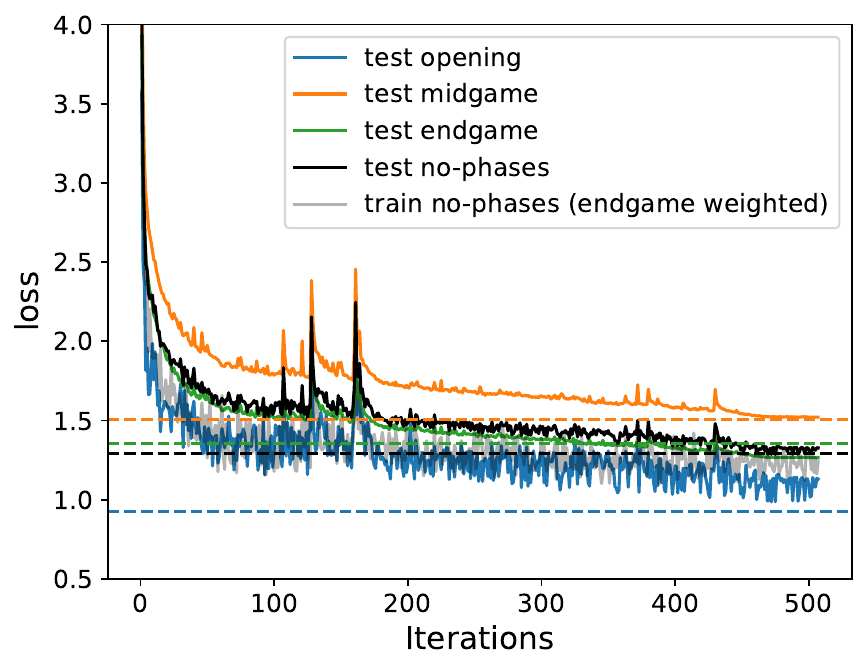}
      \label{fig:weighted_endgame_loss}
    }
    \subfloat[Policy Accuracy ($a=10$)]{\centering
      \includegraphics[width=.48\textwidth,trim=0 0 0 0,clip]{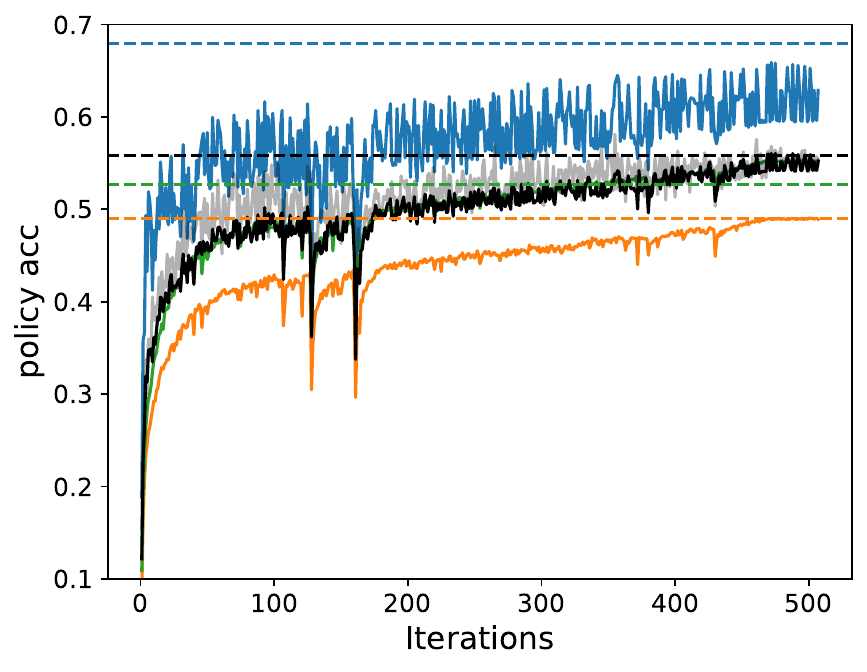}
      \label{fig:weighted_endgame_policy}
    }
    \caption{The \textit{endgame expert's} loss (Fig. \ref{fig:weighted025_endgame_loss} and Fig. \ref{fig:weighted_endgame_loss}) and policy accuracy (Fig. \ref{fig:weighted025_endgame_policy} and \ref{fig:weighted_endgame_policy}) over the course of the training process for the \textit{Weighted Learning} approach with different $a$ values. The metrics were evaluated on the unweighted opening (blue), midgame (orange), endgame (green) and no-phases (black) test set. The evaluation on the train set (weighted) is shown in the background. The values on the x-axis represent iterations, where one iteration is defined as doing backpropagation on one batch of training data. The metric values of the final model checkpoint of the Regular Learning approach are added as dashed reference lines (same colors represent the same test set).}
    \label{fig:weighted_metrics_endgame}
\end{figure*}

\paragraph{Training Process.}
From Figure \ref{fig:sep_metrics_no-phases} until Figure \ref{fig:weighted_metrics_endgame} we show the training process of our three learning approaches for all three experts on all four test sets (opening, midgame, endgame and no phases). It can be seen that the experts outperform other models in their designated phase. 

\begin{figure*}[htbp]
    \centering
    \subfloat[Elo gain by Nodes]{\centering
      \includegraphics[width=.48\textwidth,trim=0 0 0 0,clip]{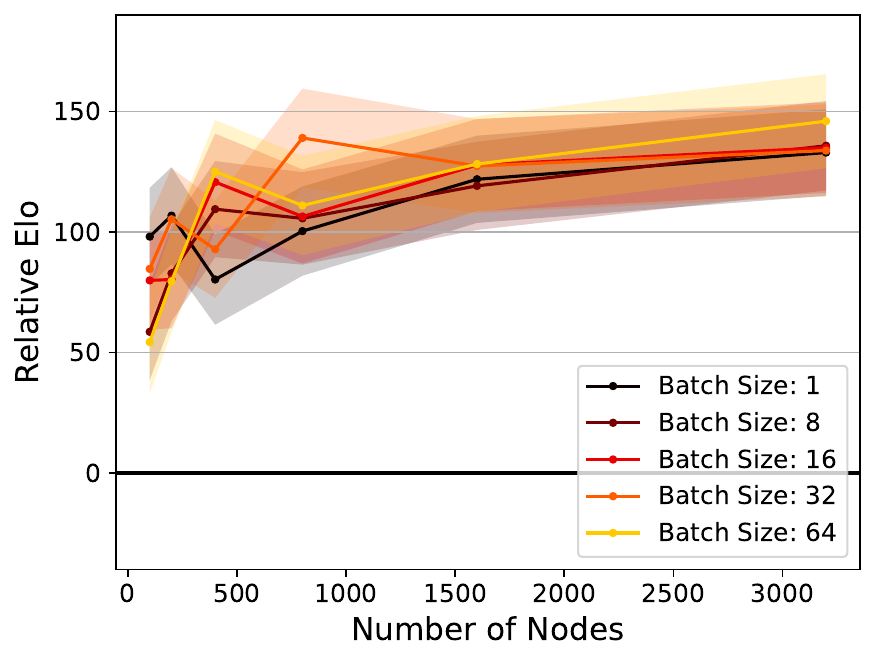}
      \label{fig:cont_nodes}
    }
    \subfloat[Elo gain by move time]{\centering
      \includegraphics[width=.48\textwidth,trim=0 0 0 0,clip]{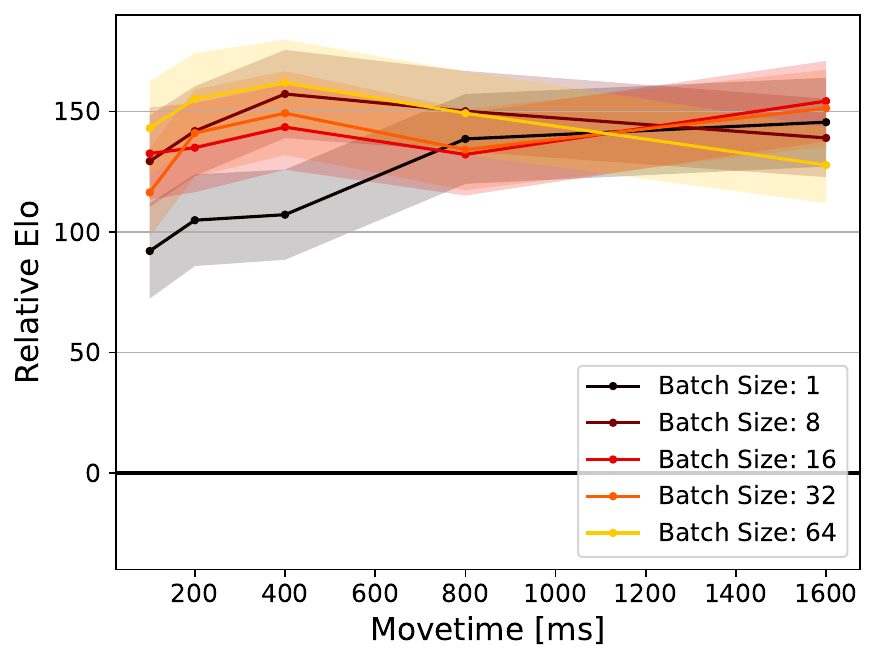}
      \label{fig:cont_movetime}
    }
    \caption{Relative Elo gain of the \textit{Staged Learning} approach for different batch sizes. Node values of 100, 200, 400, 800, 1600 and 3200 in Fig. \ref{fig:cont_nodes}. Move time values of 100, 200, 400, 800 and 1600 in Fig. \ref{fig:cont_movetime}}
    \label{fig:cont_elo}
\end{figure*}

\begin{figure*}[htbp]
    \centering
    \subfloat[Elo gain by Nodes ($a=4$)]{\centering
      \includegraphics[width=.48\textwidth,trim=0 0 0 0,clip]{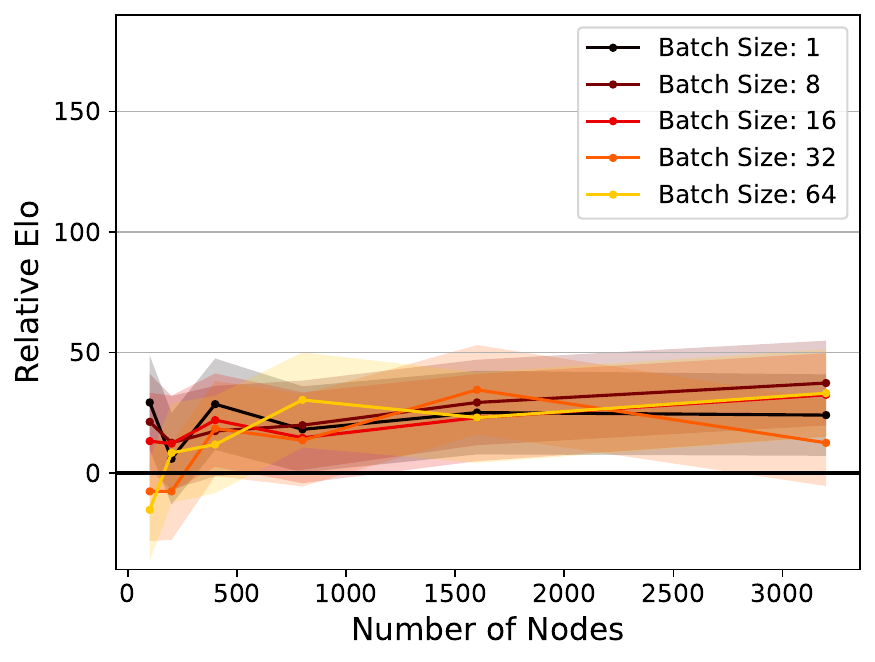}
      \label{fig:weighted_nodes}
    }
    \subfloat[Elo gain by move time ($a=4$)]{\centering
      \includegraphics[width=.48\textwidth,trim=0 0 0 0,clip]{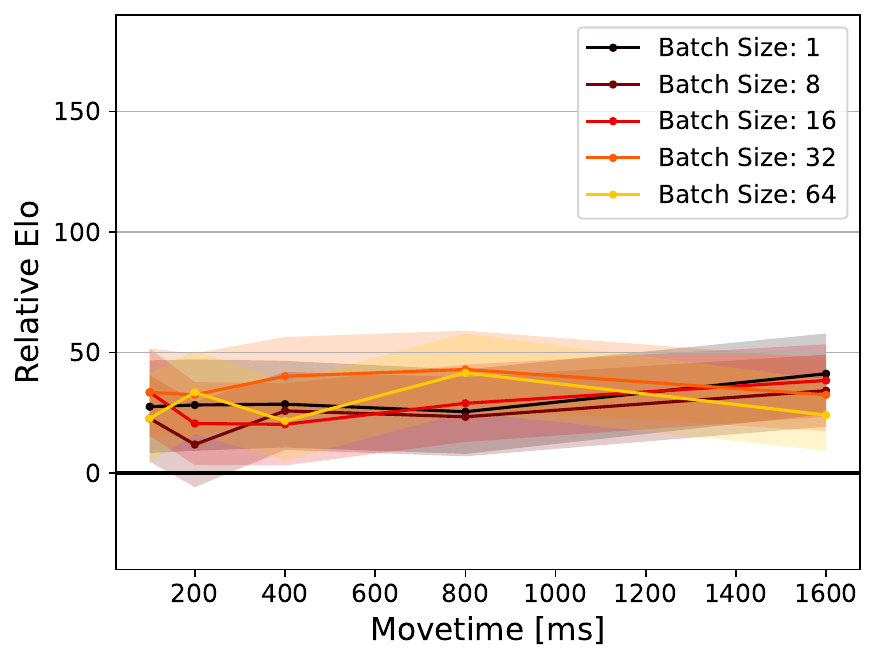}
      \label{fig:weighted_movetime}
    }\\
    \subfloat[Elo gain by Nodes ($a=10$)]{\centering
      \includegraphics[width=.48\textwidth,trim=0 0 0 0,clip]{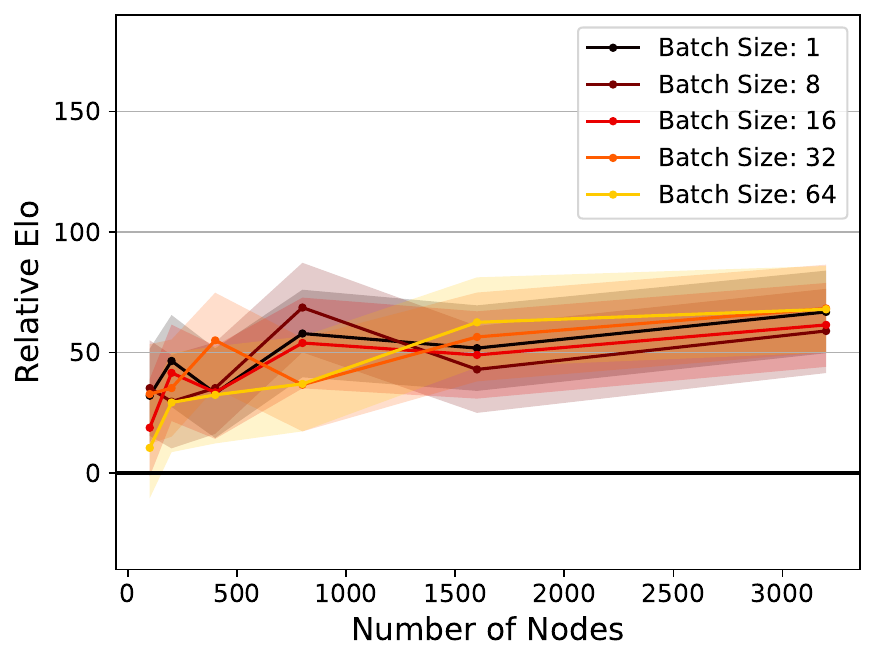}
      \label{fig:weighted01_nodes}
    }
    \subfloat[Elo gain by move time ($a=10$)]{\centering
      \includegraphics[width=.48\textwidth,trim=0 0 0 0,clip]{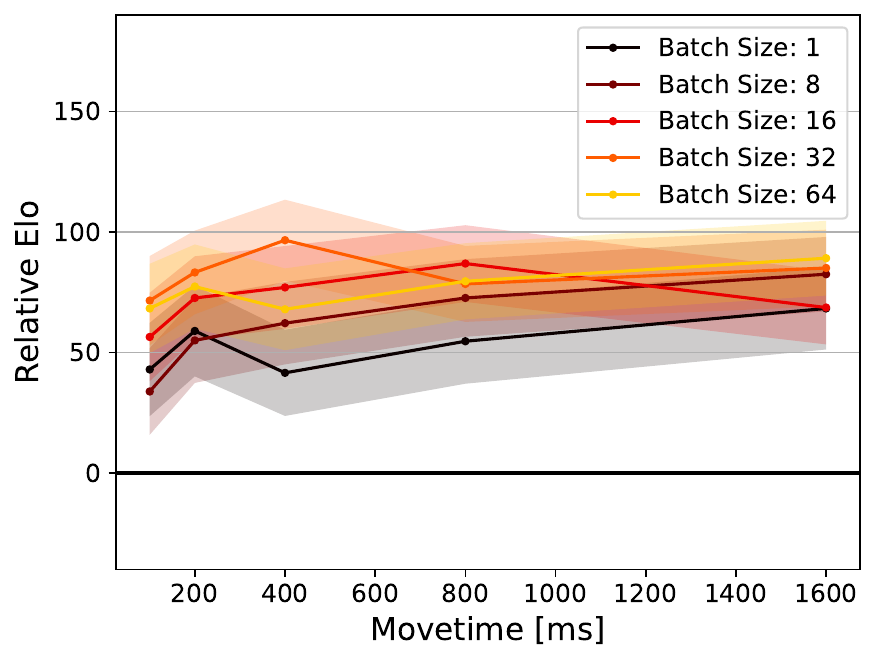}
      \label{fig:weighted01_movetime}
    }
    \caption{Relative Elo gain of the \textit{Weighted Learning} approach for different batch sizes and $a$ values of 4 and 10. Node values of 100, 200, 400, 800, 1600 and 3200 in Fig. \ref{fig:weighted_nodes} and Fig. \ref{fig:weighted01_nodes}. Move time values of 100, 200, 400, 800 and 1600 in Fig. \ref{fig:weighted_movetime} and \ref{fig:weighted01_movetime}}
    \label{fig:weighted_elo}
\end{figure*}

\paragraph{Elo Development over Training.} Figures \ref{fig:sep_elo}, \ref{fig:cont_elo} and \ref{fig:weighted_elo} show the result of test matches between the model in training and a baseline model for different restricting factors in the search and evaluation of game states. On the left we used the number of iterations, i.e., the number of explored nodes within MCTS as restricting factor. On the right, instead of restricting the search by the number of iterations, we chose to restrict it, using a time limit, a common pratice in, e.g., computational chess. As baseline model in all these figures the regular learning approach, i.e., the single model approach, was taken. 

\begin{figure}[htbp]
      \includegraphics[width=.48\textwidth,trim=0 0 0 0,clip]{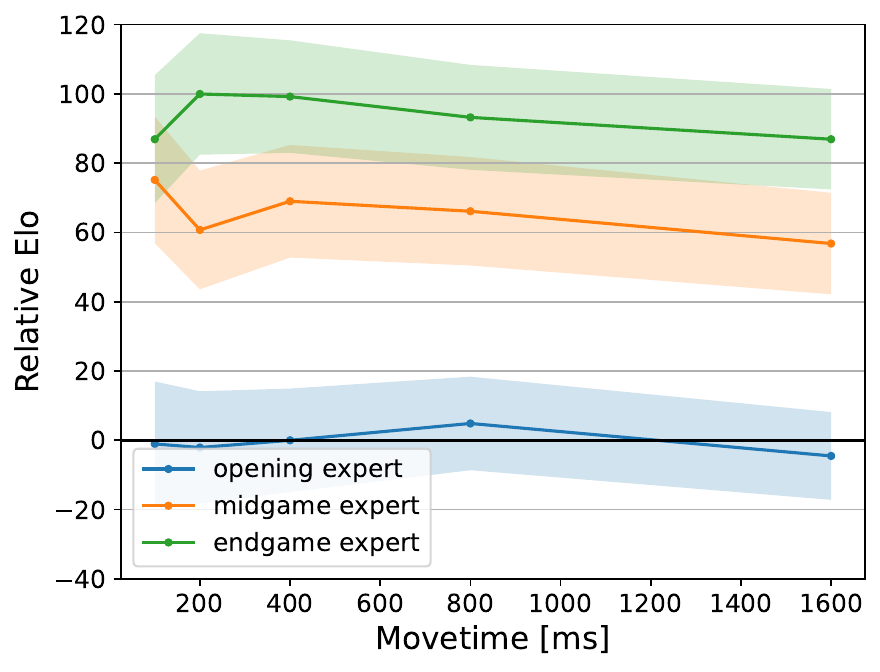}
    \caption{Relative Elo gain of using only one specific phase expert in the MCTS and using the baseline network for all remaining predictions. The used experts are taken from the separated learning approach for \textit{chess} and the batch size was set to 64 for all matches. 
    }
    \label{fig:specific_elo_movetime}
\end{figure}

\paragraph{Extension to Figure~\ref{fig:specific_elo}.} Since we did all our experiments using two restricting factors, i.e., using the number of nodes and the move time as restrictions, we also evaluated \cref{fig:specific_elo} twice. The evaluation, using time as the restricting factor, can be found in \cref{fig:specific_elo_movetime}. 

\newpage

\begin{table*}[tbh]
\centering
\resizebox{1.0\textwidth}{!}{
\begin{tabular}{@{}lclllll@{}}
\toprule
\textbf{Approach} & \textbf{Expert} & \textbf{Opening} & \textbf{Middlegame} & \textbf{Endgame} & \textbf{No-Phases} \\
\midrule
\textbf{Regular Learning} & - & \cellcolor[HTML]{FCFCFF}\textbf{0.9269} & \cellcolor[HTML]{F9F9FD}\textbf{1.5043} & \cellcolor[HTML]{FCFCFF}\textbf{1.3523} & \cellcolor[HTML]{C4D5EB}{ \textbf{1.2897}} \\
\addlinespace
\textbf{Separated Learning} & Opening & \cellcolor[HTML]{E2E9F5}\textbf{0.9169} & \cellcolor[HTML]{F8696B}2.4030 & \cellcolor[HTML]{F8696B}3.2433 & \cellcolor[HTML]{F8696B}2.2587 \\
& Middlegame & \cellcolor[HTML]{FAA2A4}1.6064 & \cellcolor[HTML]{5A8AC6}\textbf{1.4371} & \cellcolor[HTML]{FCEDF0}1.5494 & \cellcolor[HTML]{FCDDE0}1.5260 \\
& Endgame & \cellcolor[HTML]{F96C6E}2.0097 & \cellcolor[HTML]{FCE3E6}1.6605 & \cellcolor[HTML]{5A8AC6}\textbf{1.2594} & \cellcolor[HTML]{FBCDCF}1.6293 \\
& Mixture & \cellcolor[HTML]{E2E9F5}\textbf{0.9169} & \cellcolor[HTML]{5A8AC6}\textbf{1.4371} & \cellcolor[HTML]{5A8AC6}\textbf{1.2594} & \cellcolor[HTML]{5E8CC7}{ \textbf{1.2245}} \\
\addlinespace
\textbf{Staged Learning} & Opening & \cellcolor[HTML]{80A5D3}\textbf{0.8786} & \cellcolor[HTML]{FBC0C2}1.8775 & \cellcolor[HTML]{FA9FA2}2.5501 & \cellcolor[HTML]{FAAFB2}1.8153 \\
& Middlegame & \cellcolor[HTML]{FAA3A5}1.5966 & \cellcolor[HTML]{6994CB}\textbf{1.4437} & \cellcolor[HTML]{FCEFF2}1.5293 & \cellcolor[HTML]{FCDEE1}1.5203 \\
& Endgame & \cellcolor[HTML]{F8696B}2.0281 & \cellcolor[HTML]{FCE1E4}1.6748 & \cellcolor[HTML]{79A0D1}\textbf{1.2774} & \cellcolor[HTML]{FBCACD}1.6458 \\
& Mixture & \cellcolor[HTML]{80A5D3}\textbf{0.8786} & \cellcolor[HTML]{6994CB}\textbf{1.4437} & \cellcolor[HTML]{79A0D1}\textbf{1.2774} & \cellcolor[HTML]{5A8AC6}{ \textbf{1.2218}} \\
\addlinespace
\textbf{Weighted Learning ($a=4$)} & Opening & \cellcolor[HTML]{5A8AC6}\textbf{0.8636} & \cellcolor[HTML]{FCFBFE}1.5125 & \cellcolor[HTML]{FCFBFD}1.3780 & \cellcolor[HTML]{B8CCE7}1.2820 \\
& Middlegame & \cellcolor[HTML]{FCF7FA}0.9661 & \cellcolor[HTML]{FCFCFF}1.5055 & \cellcolor[HTML]{FCF7FA}1.4207 & \cellcolor[HTML]{FCFCFF}1.3248 \\
& Endgame & \cellcolor[HTML]{FCFCFF}0.9339 & \cellcolor[HTML]{F8F9FD}\textbf{1.5041} & \cellcolor[HTML]{89ABD6}\textbf{1.2864} & \cellcolor[HTML]{A4BEE0}1.2695 \\
& Mixture & \cellcolor[HTML]{5A8AC6}\textbf{0.8636} & \cellcolor[HTML]{FCFCFF}1.5055 & \cellcolor[HTML]{89ABD6}\textbf{1.2864} & \cellcolor[HTML]{7BA1D1}{ \textbf{1.2433}} \\
\addlinespace
\textbf{Weighted Learning ($a=10$)} & Opening & \cellcolor[HTML]{9AB7DC}\textbf{0.8887} & \cellcolor[HTML]{FCE3E6}1.6600 & \cellcolor[HTML]{FCEEF1}1.5389 & \cellcolor[HTML]{FCF1F4}1.3995 \\
& Middlegame & \cellcolor[HTML]{FCF3F6}0.9955 & \cellcolor[HTML]{C7D7EC}\textbf{1.4834} & \cellcolor[HTML]{FCF5F8}1.4470 & \cellcolor[HTML]{FCFBFE}1.3346 \\
& Endgame & \cellcolor[HTML]{FCEEF1}1.0334 & \cellcolor[HTML]{FCF4F7}1.5547 & \cellcolor[HTML]{A7C0E1}\textbf{1.3036} & \cellcolor[HTML]{FAFBFE}1.3241 \\
& Mixture & \cellcolor[HTML]{9AB7DC}\textbf{0.8887} & \cellcolor[HTML]{C7D7EC}\textbf{1.4834} & \cellcolor[HTML]{A7C0E1}\textbf{1.3036} & \cellcolor[HTML]{83A7D4}{ \textbf{1.2483}} \\
\bottomrule
\end{tabular}}
\caption{{MoE reduces overall loss in chess using game phase specific experts.} This table provides a comprehensive overview of the loss values for each expert's final model across different approaches. Performance is evaluated over four different test sets: opening, middlegame, endgame, and no-phases. In addition to the individual expert losses, we include the loss for the resulting mixture model that includes all three experts. This mixture model determines the appropriate expert based on the phase of the current position in the test set. The weighted learning approach was analyzed with different weighting factors to emphasize the main phase over others.
The table uses color coding for each column, ranging from high loss (red) to low loss (blue). Bold text highlights the most effective model within each learning process for each test set, while the best performing model in the no-phases test set is underlined, highlighting its overall superiority. This is a more detailed version of \cref{tab:short_loss}.}
\label{tab:loss_overview}
\end{table*}

\begin{table*}[t]
\resizebox{\linewidth}{!}{
\begin{tabular}{@{}c|cccc} \toprule
\multicolumn{1}{l|}{\textbf{Batchsize}}& \multicolumn{1}{l}{\textbf{Separated Learning}} & \multicolumn{1}{l}{\textbf{Staged Learning}} & \multicolumn{1}{l}{\textbf{Weighted Learning}} & \multicolumn{1}{l}{\textbf{Weighted Learning}} \\
\multicolumn{1}{l|}{} & \multicolumn{2}{l}{} & $a=4$ & $a=10$ \\\midrule
\textbf{1}       & 106.89$\pm$18.29                               & 111.68$\pm$20.50                                       & \textbf{25.61}$\pm$8.55                                         & 50.41$\pm$12.33                                                   \\
\textbf{8}       & 123.45$\pm$18.57                               & 120.81$\pm$29.89                                       & 23.20$\pm$8.10                                                  & 52.42$\pm$18.07                                                   \\
\textbf{16}      & 126.81$\pm$20.56                               & 122.49$\pm$24.17                                       & 23.52$\pm$8.75                                                  & 56.33$\pm$20.04                                                   \\
\textbf{32}      & 124.36$\pm$19.25                               & 125.00$\pm$22.42                                       & 22.28$\pm$17.90                                                  & \textbf{63.55}$\pm$22.01                                          \\
\textbf{64}      & \textbf{129.49$\pm31.21$} & \textbf{125.55}$\pm$33.02                              & 21.31$\pm$15.45                                                  & 56.51$\pm$25.08                                                   \\
\textbf{Average} & 122.20$\pm22.54$                               & 121.11$\pm$25.68                                       & 23.18$\pm$11.95                                                 & 55.84$\pm$19.57  \\\bottomrule                                             
\end{tabular}}
\caption{{M2CTS outperform our baseline ``one-for-all'' MCTS approach in direct comparison.} This table presents the relative Elo gains achieved by our different training approaches over our baseline model across different batch sizes. A higher Elo value correspond to a higher difference in playing strength in favor of M2CTS over MCTS. The highest Elo gain is marked \textbf{bold}. This is a more detailed version of \cref{tab:elo_overview}.}
\label{app:m2ctsvsmcts}
\end{table*}

\begin{figure*}[tbh]
    \centering
    \subfloat[Loss]{\centering
      \includegraphics[width=.48\textwidth,trim=0 0 0 0,clip]{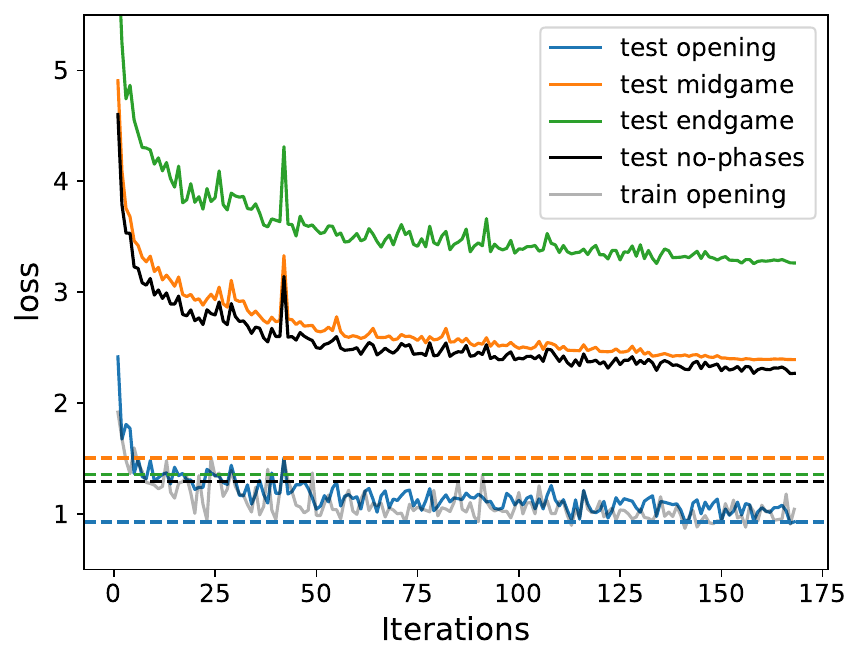}
      \label{fig:correct_opening_loss}
    }
    \subfloat[Policy Accuracy]{\centering
      \includegraphics[width=.48\textwidth,trim=0 0 0 0,clip]{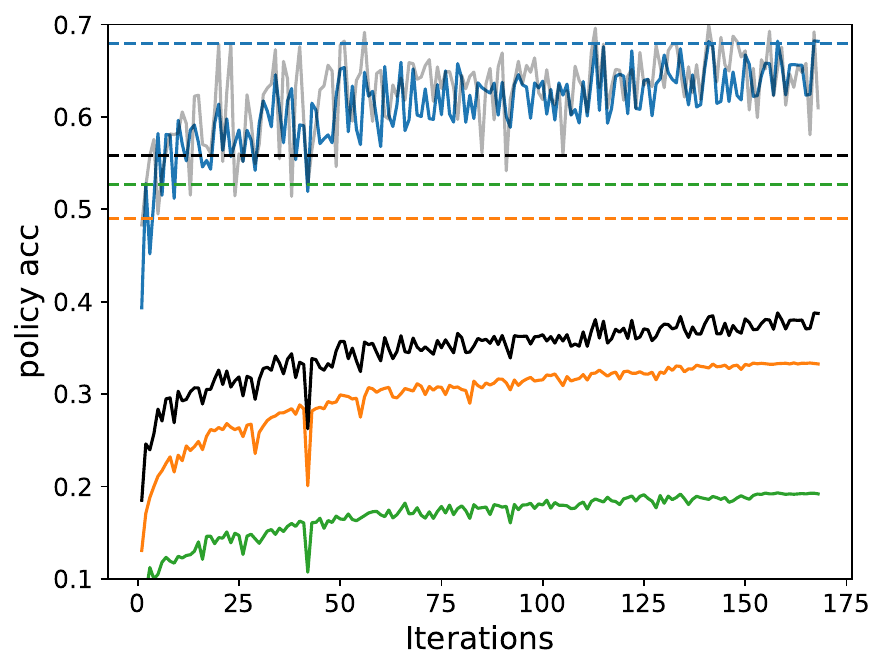}
      \label{fig:correct_opening_policy}
    }
    \caption{The \textit{opening expert's} loss (Fig. \ref{fig:correct_opening_loss}) and policy accuracy (Fig. \ref{fig:correct_opening_policy}) over the course of the training process for the \textit{Separated Learning} approach. The metrics were evaluated on the opening (blue), midgame (orange), endgame (green) and no-phases (black) test set. The evaluation on the train set is shown in the background in gray. The values on the x-axis represent iterations, where one iteration is defined as doing backpropagation on one batch of training data.  The metric values of the final model checkpoint of the Regular Learning approach are added as dashed reference lines (same colors represent the same test set).}
    \label{fig:sep_metrics_opening}
\end{figure*}

\begin{figure*}[tbh]
    \centering
    \subfloat[Loss]{\centering
      \includegraphics[width=.48\textwidth,trim=0 0 0 0,clip]{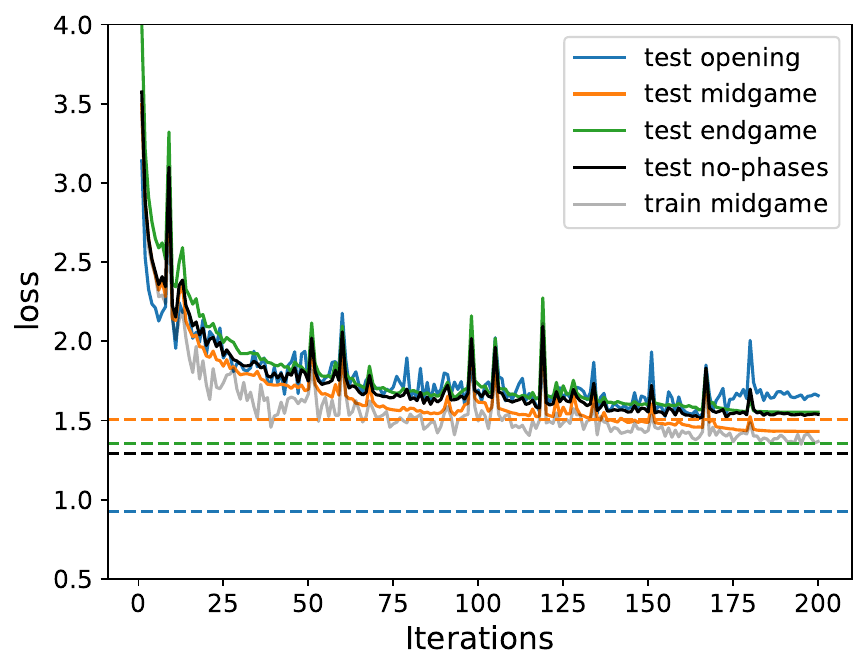}
      \label{fig:correct_midgame_loss}
    }
    \subfloat[Policy Accuracy]{\centering
      \includegraphics[width=.48\textwidth,trim=0 0 0 0,clip]{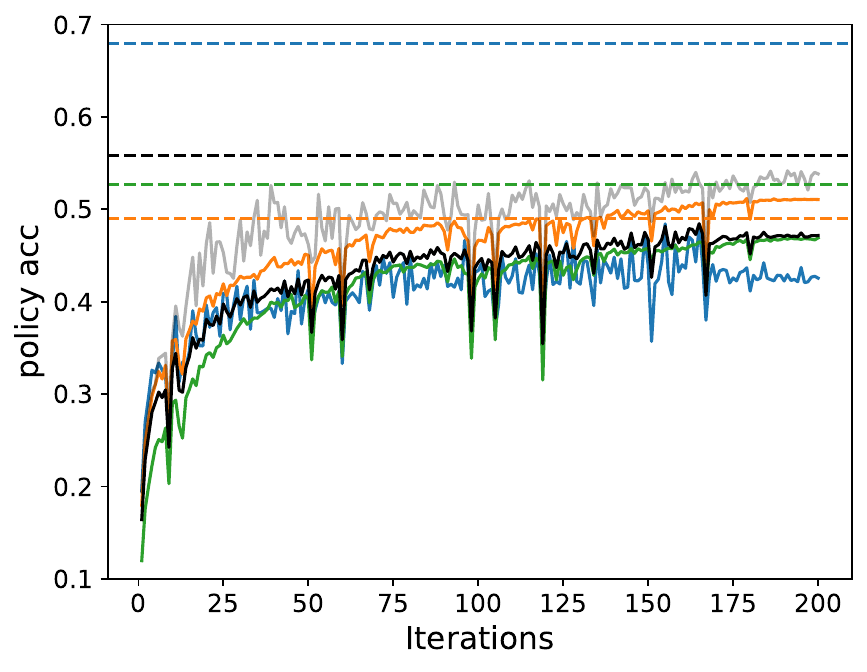}
      \label{fig:correct_midgame_policy}
    }
    \caption{The \textit{midgame expert's} loss (Fig. \ref{fig:correct_midgame_loss}) and policy accuracy (Fig. \ref{fig:correct_midgame_policy}) over the course of the training process for the \textit{Separated Learning} approach. The metrics were evaluated on the opening (blue), midgame (orange), endgame (green) and no-phases (black) test set. The evaluation on the train set is shown in the background in gray. The values on the x-axis represent iterations, where one iteration is defined as doing backpropagation on one batch of training data.  The metric values of the final model checkpoint of the Regular Learning approach are added as dashed reference lines (same colors represent the same test set).}
    \label{fig:sep_metrics_midgame}
\end{figure*}

\begin{figure*}[tbh]
    \centering
    \subfloat[Loss]{\centering
      \includegraphics[width=.48\textwidth,trim=0 0 0 0,clip]{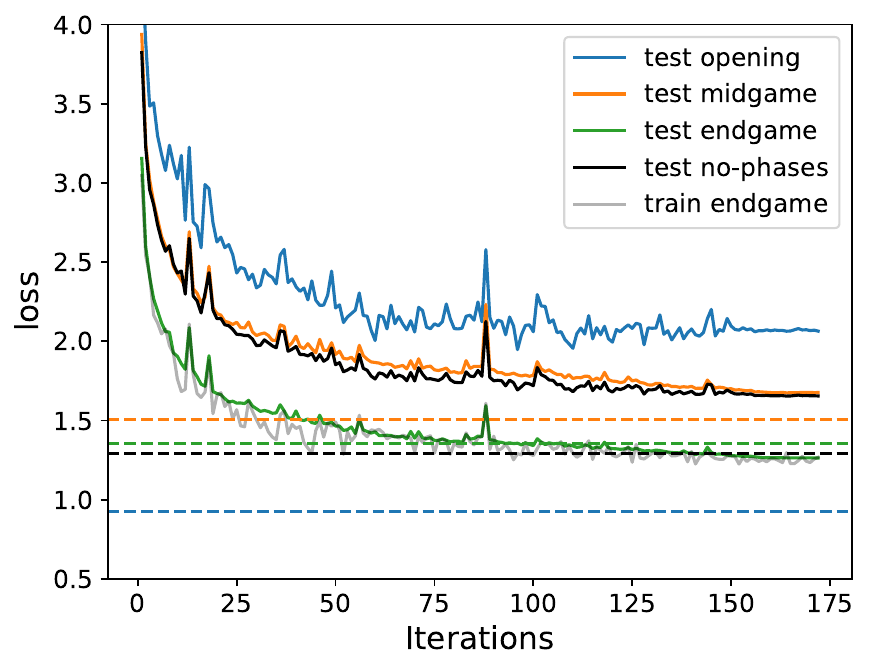}
      \label{fig:correct_endgame_loss}
    }
    \subfloat[Policy Accuracy]{\centering
      \includegraphics[width=.48\textwidth,trim=0 0 0 0,clip]{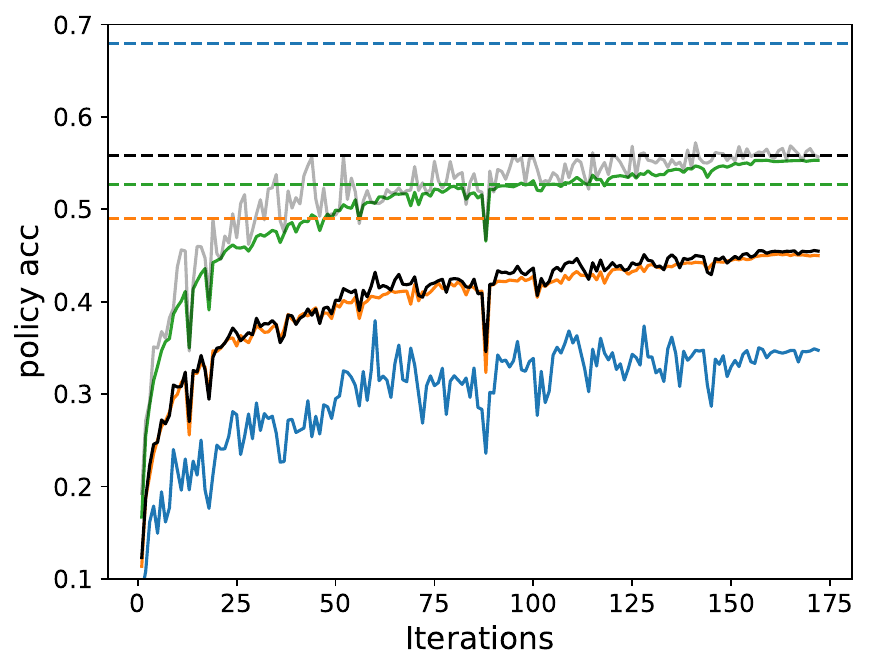}
      \label{fig:correct_endgame_policy}
    }
    \caption{The \textit{endgame expert's} loss (Fig. \ref{fig:correct_endgame_loss}) and policy accuracy (Fig. \ref{fig:correct_endgame_policy}) over the course of the training process for the \textit{Separated Learning} approach. The metrics were evaluated on the opening (blue), midgame (orange), endgame (green) and no-phases (black) test set. The evaluation on the train set is shown in the background in gray. The values on the x-axis represent iterations, where one iteration is defined as doing backpropagation on one batch of training data.  The metric values of the final model checkpoint of the Regular Learning approach are added as dashed reference lines (same colors represent the same test set).}
    \label{fig:sep_metrics_endgame}
\end{figure*}

\begin{figure*}[tbh]
    \centering
    \subfloat[Loss]{\centering
      \includegraphics[width=.48\textwidth,trim=0 0 0 0,clip]{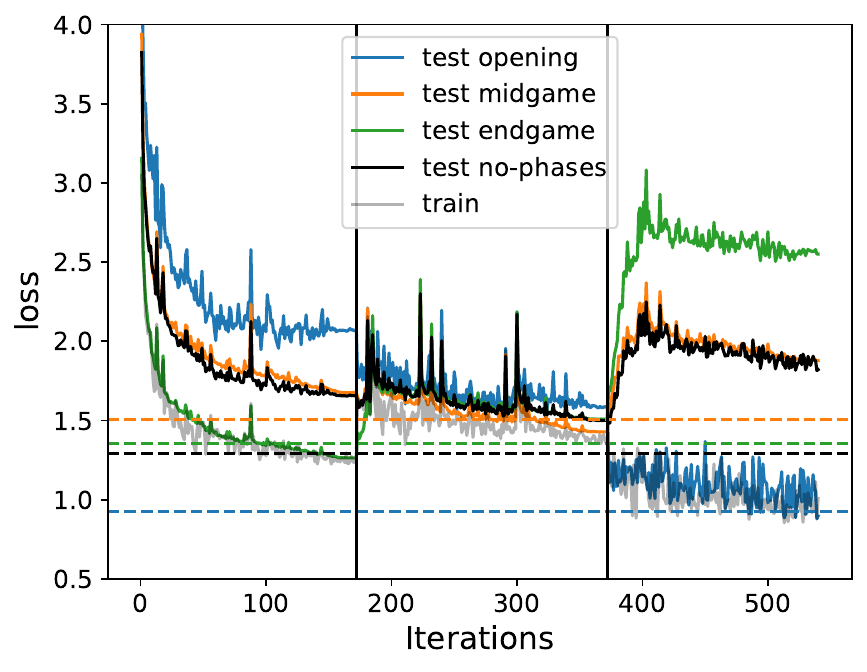}
      \label{fig:cont_opening_loss}
    }
    \subfloat[Policy Accuracy]{\centering
      \includegraphics[width=.48\textwidth,trim=0 0 0 0,clip]{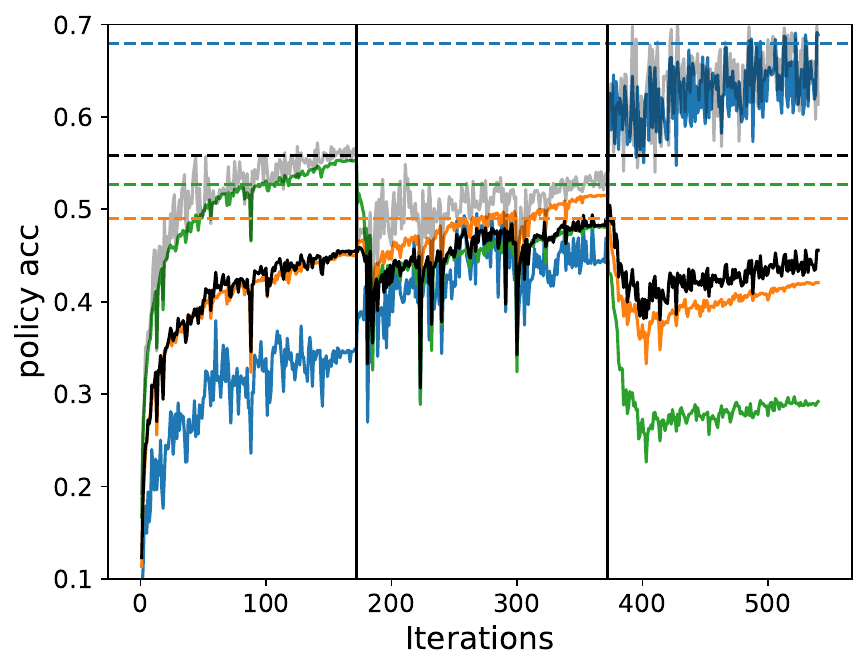}
      \label{fig:cont_opening_policy}
    }
    \caption{The \textit{opening expert's} loss (Fig. \ref{fig:cont_opening_loss}) and policy accuracy (Fig. \ref{fig:cont_opening_policy}) over the course of the training process for the \textit{Staged Learning} approach. The metrics were evaluated on the opening (blue), midgame (orange), endgame (green) and no-phases (black) test set. The evaluation on the train set is shown in the background. The values on the x-axis represent iterations, where one iteration is defined as doing backpropagation on one batch of training data. The vertical lines indicate the points at which the training on a new dataset started (initialized with the parameters of the last model checkpoint of the previous training stage).}
    \label{fig:cont_metrics_opening}
\end{figure*}

\begin{figure*}[tbh]
    \centering
    \subfloat[Loss]{\centering
      \includegraphics[width=.48\textwidth,trim=0 0 0 0,clip]{media/supplements/cont_phases_opening_loss.pdf}
      \label{fig:cont_opening_loss}
    }
    \subfloat[Policy Accuracy]{\centering
      \includegraphics[width=.48\textwidth,trim=0 0 0 0,clip]{media/supplements/cont_phases_opening_policy_acc.pdf}
      \label{fig:cont_opening_policy}
    }
    \caption{The \textit{opening expert's} loss (Fig. \ref{fig:cont_opening_loss}) and policy accuracy (Fig. \ref{fig:cont_opening_policy}) over the course of the training process for the \textit{Staged Learning} approach. The metrics were evaluated on the opening (blue), midgame (orange), endgame (green) and no-phases (black) test set. The evaluation on the train set is shown in the background. The values on the x-axis represent iterations, where one iteration is defined as doing backpropagation on one batch of training data. The vertical lines indicate the points at which the training on a new dataset started (initialized with the parameters of the last model checkpoint of the previous training stage).}
    \label{fig:cont_metrics_opening}
\end{figure*}

\begin{figure*}[tbh]
    \centering
    \subfloat[Loss]{\centering
      \includegraphics[width=.48\textwidth,trim=0 0 0 0,clip]{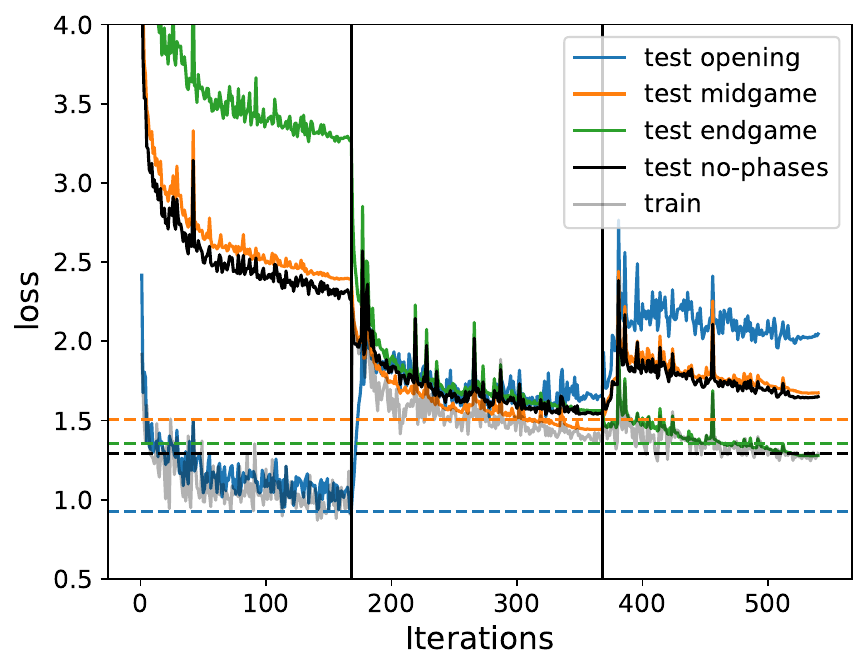}
      \label{fig:cont_endgame_loss}
    }
    \subfloat[Policy Accuracy]{\centering
      \includegraphics[width=.48\textwidth,trim=0 0 0 0,clip]{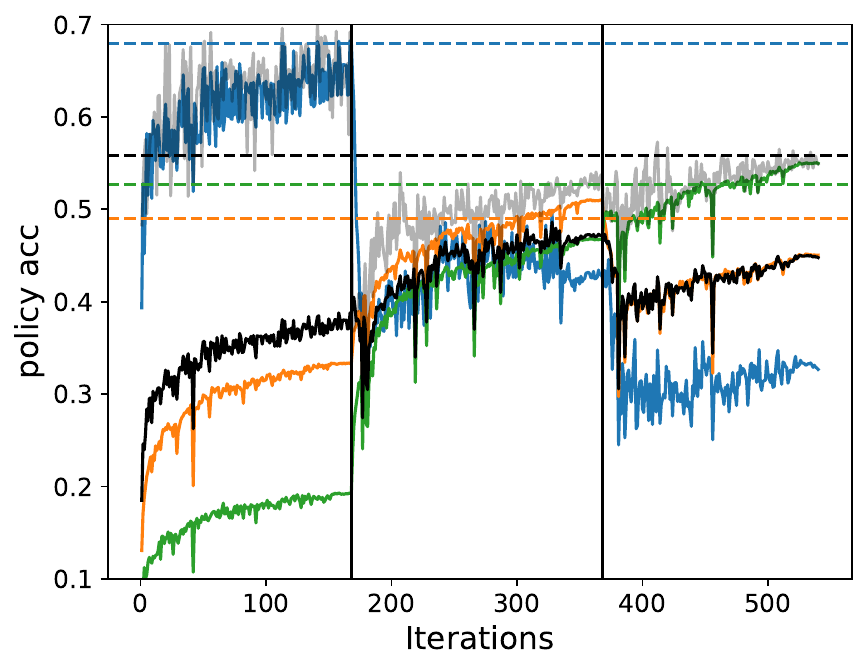}
      \label{fig:cont_endgame_policy}
    }
    \caption{The \textit{endgame expert's} loss (Fig. \ref{fig:cont_endgame_loss}) and policy accuracy (Fig. \ref{fig:cont_endgame_policy}) over the course of the training process for the \textit{Staged Learning} approach. The metrics were evaluated on the opening (blue), midgame (orange), endgame (green) and no-phases (black) test set. The evaluation on the train set is shown in the background. The values on the x-axis represent iterations, where one iteration is defined as doing backpropagation on one batch of training data. The vertical lines indicate the points at which the training on a new dataset started (initialized with the parameters of the last model checkpoint of the previous training stage).}
    \label{fig:cont_metrics_endgame}
\end{figure*}

\begin{figure*}[tbh]
    \centering
    \subfloat[Loss ($a=4$)]{\centering
      \includegraphics[width=.48\textwidth,trim=0 0 0 0,clip]{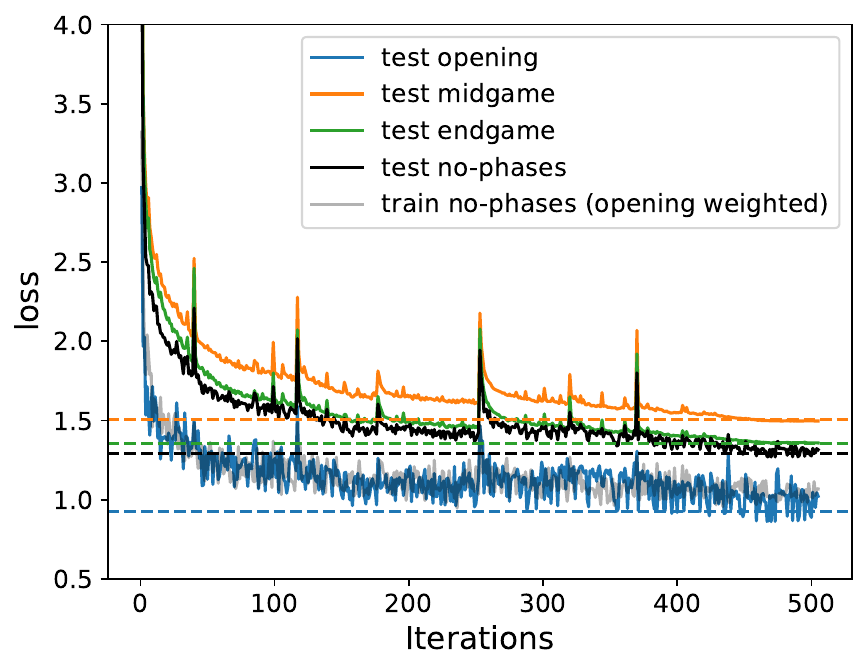}
      \label{fig:weighted025_opening_loss}
    }
    \subfloat[Policy Accuracy ($a=4$)]{\centering
      \includegraphics[width=.48\textwidth,trim=0 0 0 0,clip]{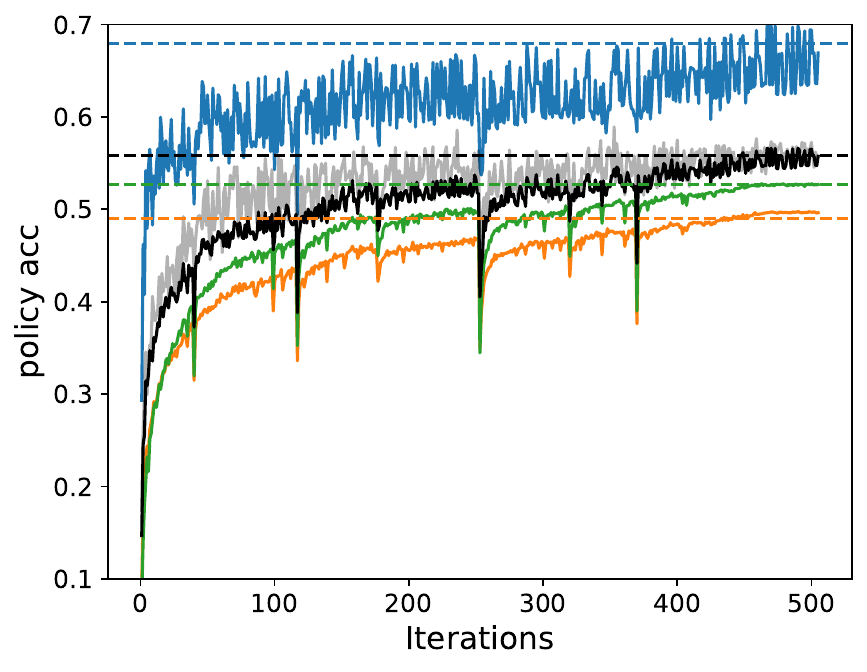}
      \label{fig:weighted025_opening_policy}
    }\\
    
    \subfloat[Loss ($a=10$)]{\centering
      \includegraphics[width=.48\textwidth,trim=0 0 0 0,clip]{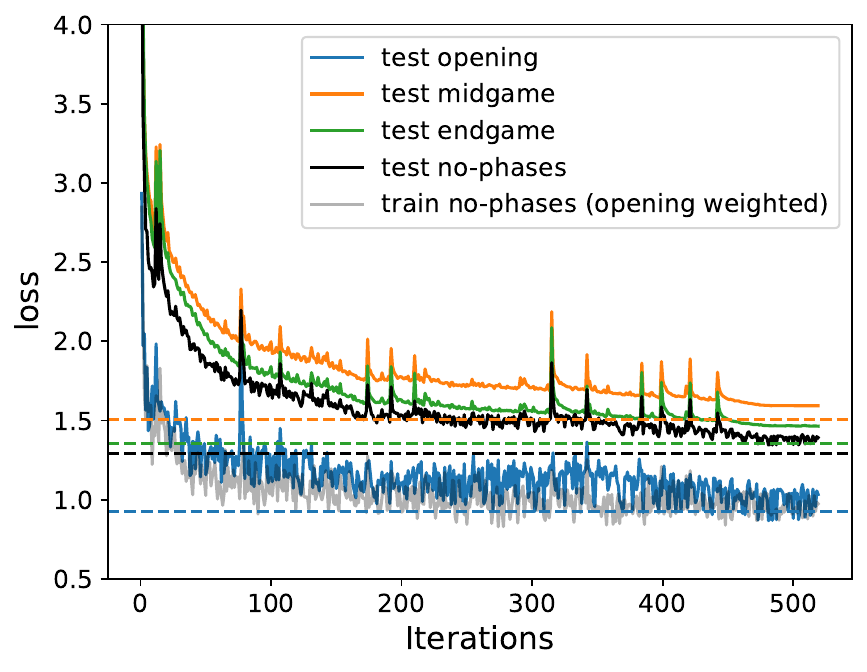}
      \label{fig:weighted_opening_loss}
    }
    \subfloat[Policy Accuracy ($a=10$)]{\centering
      \includegraphics[width=.48\textwidth,trim=0 0 0 0,clip]{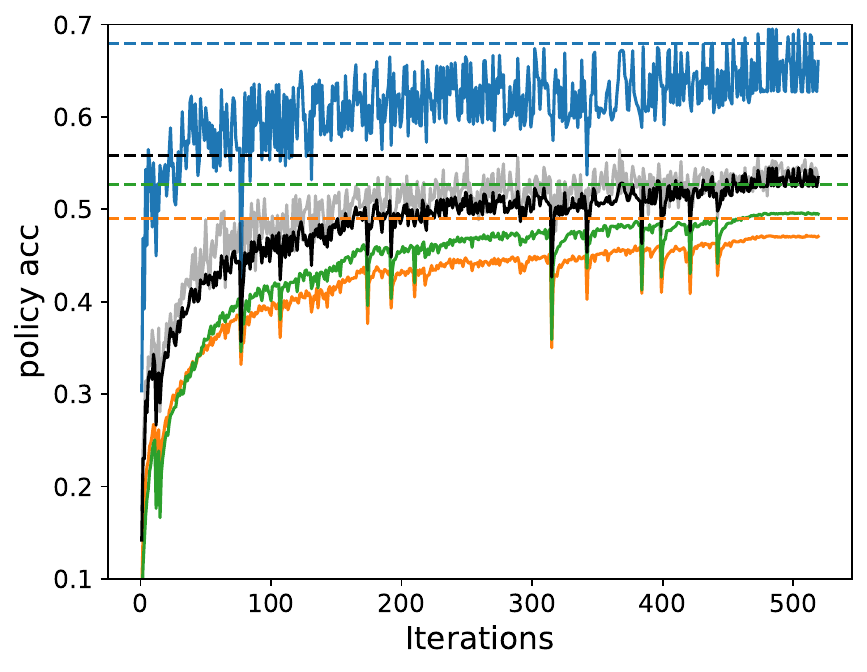}
      \label{fig:weighted_opening_policy}
    }

    \caption{The \textit{opening expert's} loss (Fig. \ref{fig:weighted025_opening_loss} and Fig. \ref{fig:weighted_opening_loss}) and policy accuracy (Fig. \ref{fig:weighted025_opening_policy} and Fig. \ref{fig:weighted_opening_policy}) over the course of the training process for the \textit{Weighted Learning} approach with different $a$ values. The metrics were evaluated on the unweighted opening (blue), midgame (orange), endgame (green) and no-phases (black) test set. The evaluation on the train set (weighted) is shown in the background. The values on the x-axis represent iterations, where one iteration is defined as doing backpropagation on one batch of training data. The metric values of the final model checkpoint of the Regular Learning approach are added as dashed reference lines (same colors represent the same test set).}
    \label{fig:weighted_metrics_opening}
\end{figure*}

\begin{figure*}[tbh]
    \centering
        \subfloat[Loss ($a=4$)]{\centering
      \includegraphics[width=.48\textwidth,trim=0 0 0 0,clip]{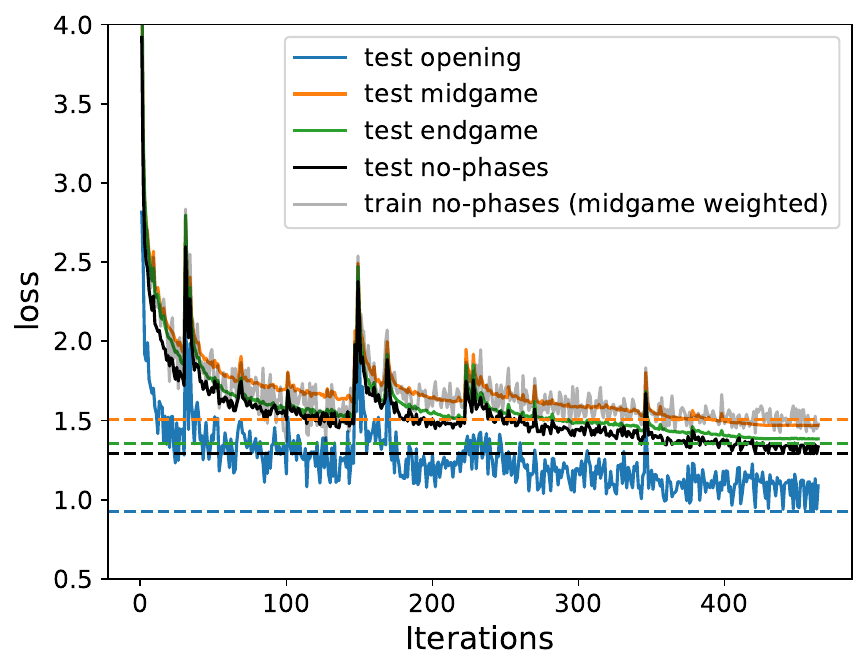}
      \label{fig:weighted025_midgame_loss}
    }
    \subfloat[Policy Accuracy ($a=4$)]{\centering
      \includegraphics[width=.48\textwidth,trim=0 0 0 0,clip]{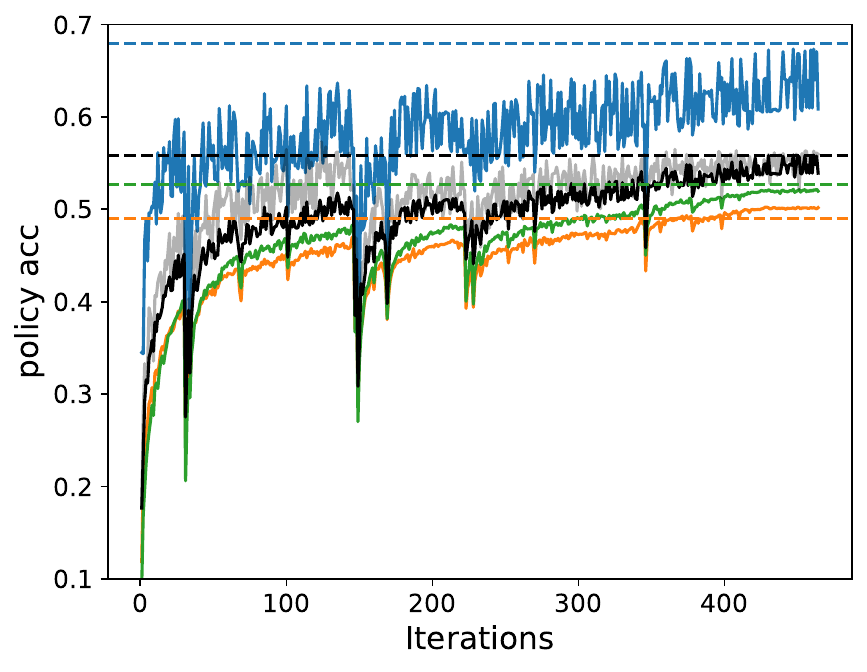}
      \label{fig:weighted025_midgame_policy}
    }\\
    \subfloat[Loss ($a=10$)]{\centering
      \includegraphics[width=.48\textwidth,trim=0 0 0 0,clip]{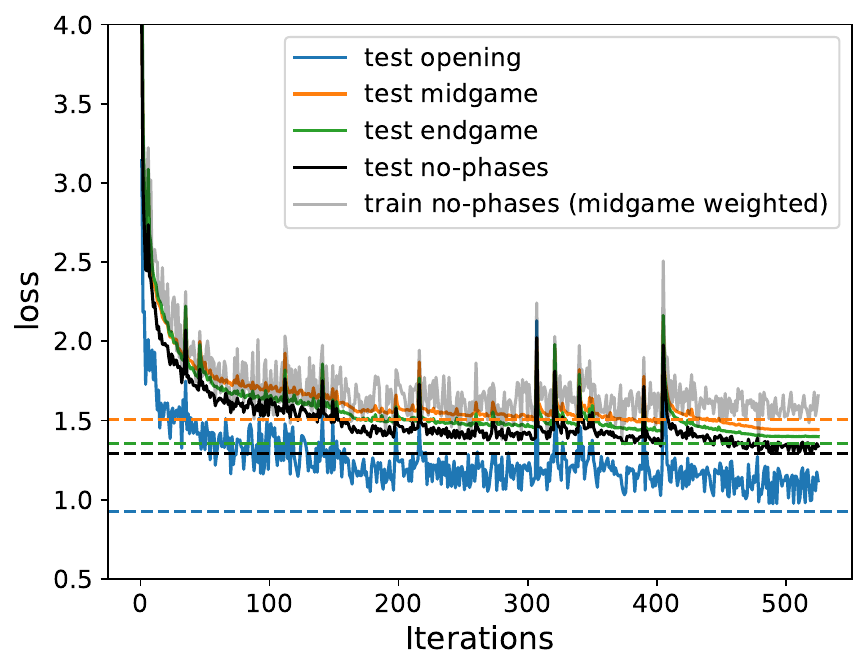}
      \label{fig:weighted_midgame_loss}
    }
    \subfloat[Policy Accuracy ($a=10$)]{\centering
      \includegraphics[width=.48\textwidth,trim=0 0 0 0,clip]{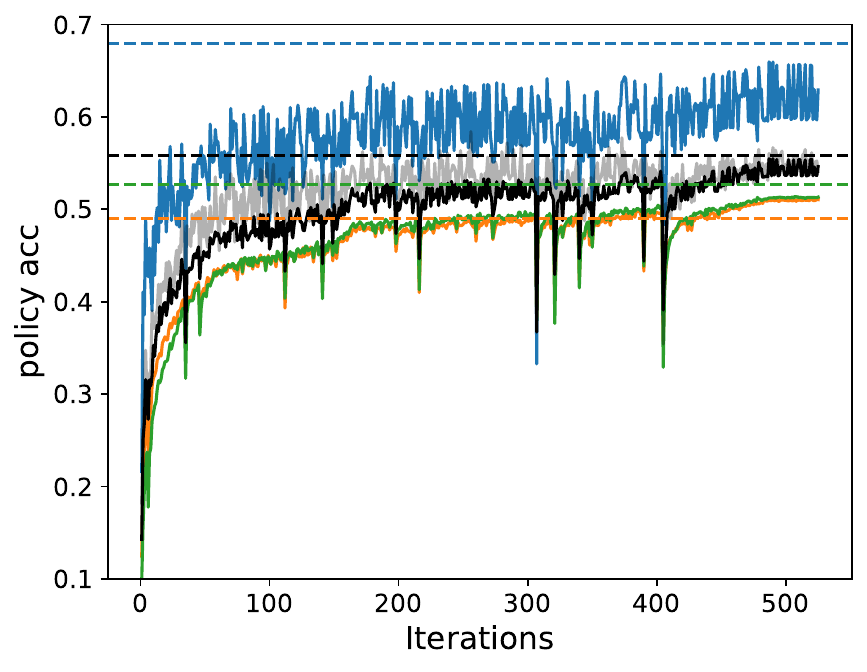}
      \label{fig:weighted_midgame_policy}
    }
    \caption{The \textit{midgame expert's} loss (Fig. \ref{fig:weighted025_midgame_loss} and Fig. \ref{fig:weighted_midgame_loss}) and policy accuracy (Fig. \ref{fig:weighted025_midgame_policy} and Fig. \ref{fig:weighted_midgame_policy}) over the course of the training process for the \textit{Weighted Learning} approach with different $a$ values. The metrics were evaluated on the unweighted opening (blue), midgame (orange), endgame (green) and no-phases (black) test set. The evaluation on the train set (weighted) is shown in the background. The values on the x-axis represent iterations, where one iteration is defined as doing backpropagation on one batch of training data. The metric values of the final model checkpoint of the Regular Learning approach are added as dashed reference lines (same colors represent the same test set).}
    \label{fig:weighted_metrics_midgame}
\end{figure*}

\begin{figure*}[tbh]
    \centering
    \subfloat[Elo gain by Nodes]{\centering
      \includegraphics[width=.48\textwidth,trim=0 0 0 0,clip]{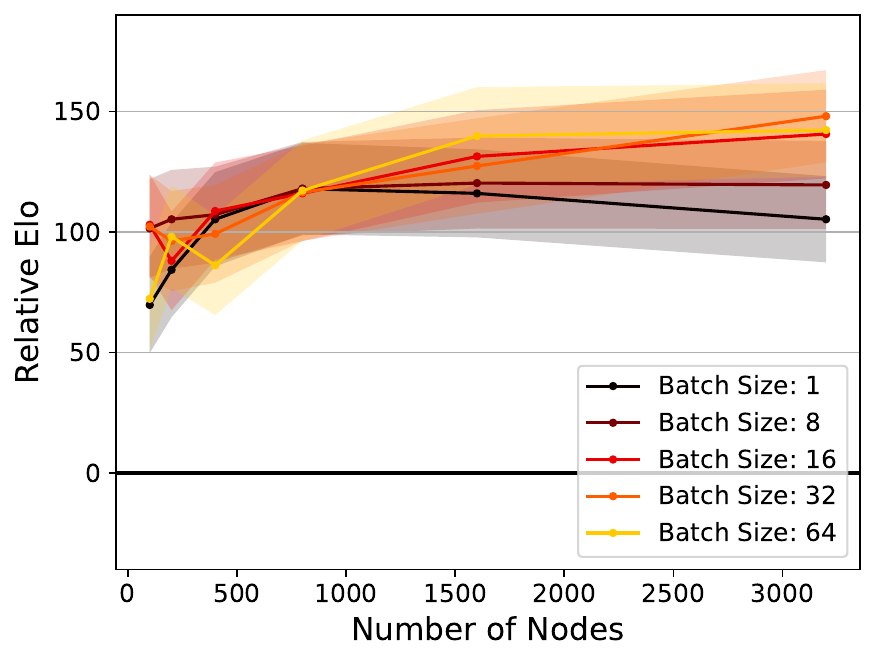}
      \label{fig:sep_nodes}
    }
    \subfloat[Elo gain by move time]{\centering
      \includegraphics[width=.48\textwidth,trim=0 0 0 0,clip]{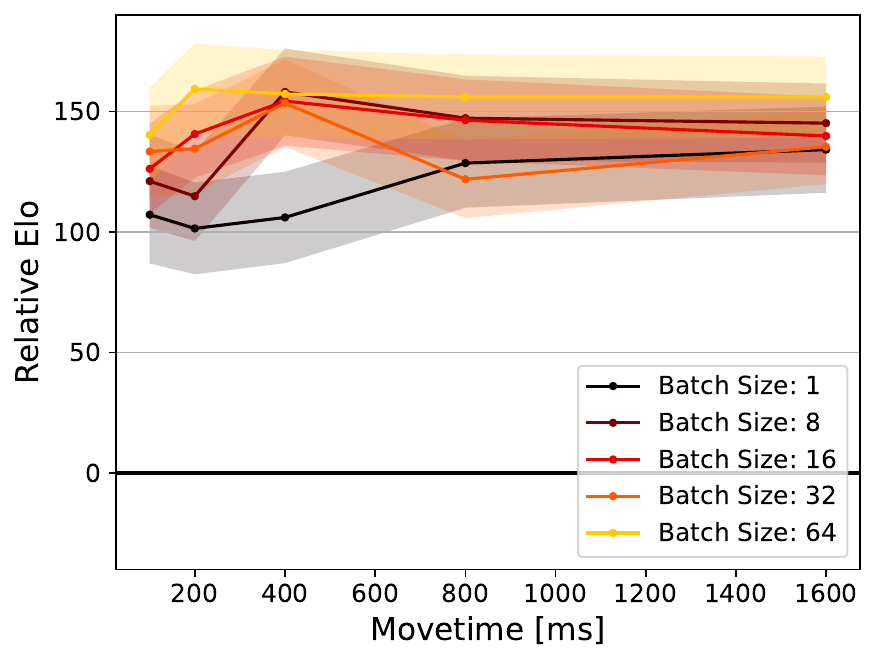}
      \label{fig:sep_movetime}
    }
    \caption{Relative Elo gain of the \textit{Separated Learning} approach for different batch sizes. Node values of 100, 200, 400, 800, 1600 and 3200 in Fig. \ref{fig:sep_nodes}. Move time values of 100, 200, 400, 800 and 1600 in Fig. \ref{fig:sep_movetime}}
    \label{app:sep_elo}
\end{figure*}

\clearpage
\newpage

\end{document}